\documentclass[11pt,a4paper,twoside]{book}
\usepackage[utf8]{inputenc}
\title{Graduation Project}
\author{Mohammad Al-Fetyani}
\date{November 2019}

\usepackage{geometry}
\usepackage{natbib}
\usepackage{graphicx}
\usepackage{float}
\usepackage{amsmath}
\usepackage{amssymb}
\usepackage{tikz}
\usepackage{pdfpages}
\usepackage{bm}
\usepackage{hyperref}
\usepackage{booktabs}
\usepackage{tabularx}
\usepackage{subcaption}
\usepackage{pgfgantt}
\usepackage{pgfplots} 
\usepackage{grffile}
\usepackage{epsfig}
\graphicspath{{./MatlabPlots/}{./Figures/}}
\usepackage{epstopdf}

\pgfplotsset{compat=newest}
\usetikzlibrary{plotmarks}
\usetikzlibrary{arrows.meta}
\usepgfplotslibrary{patchplots}
\newlength\fwidth
\usepackage{color, colortbl}
\definecolor{Gray}{gray}{0.9}
\newcommand*{\belowrulesepcolor}[1]{%
  \noalign{%
    \kern-\belowrulesep
    \begingroup
      \color{#1}%
      \hrule height\belowrulesep
    \endgroup
  }%
}
\newcommand*{\aboverulesepcolor}[1]{%
  \noalign{%
    \begingroup
      \color{#1}%
      \hrule height\aboverulesep
    \endgroup
    \kern-\aboverulesep
  }%
}

\usepackage{caption} 
\usetikzlibrary{shapes,arrows,fit,calc,positioning,automata}

\newenvironment{bnmatrix}
    {\left[\;\begin{matrix}
    }
    {
    \end{matrix}\;\right]
    }

\setcitestyle{square,numbers,sort,comma,compress}

\begin{document}
\includepdf[pages={1}]{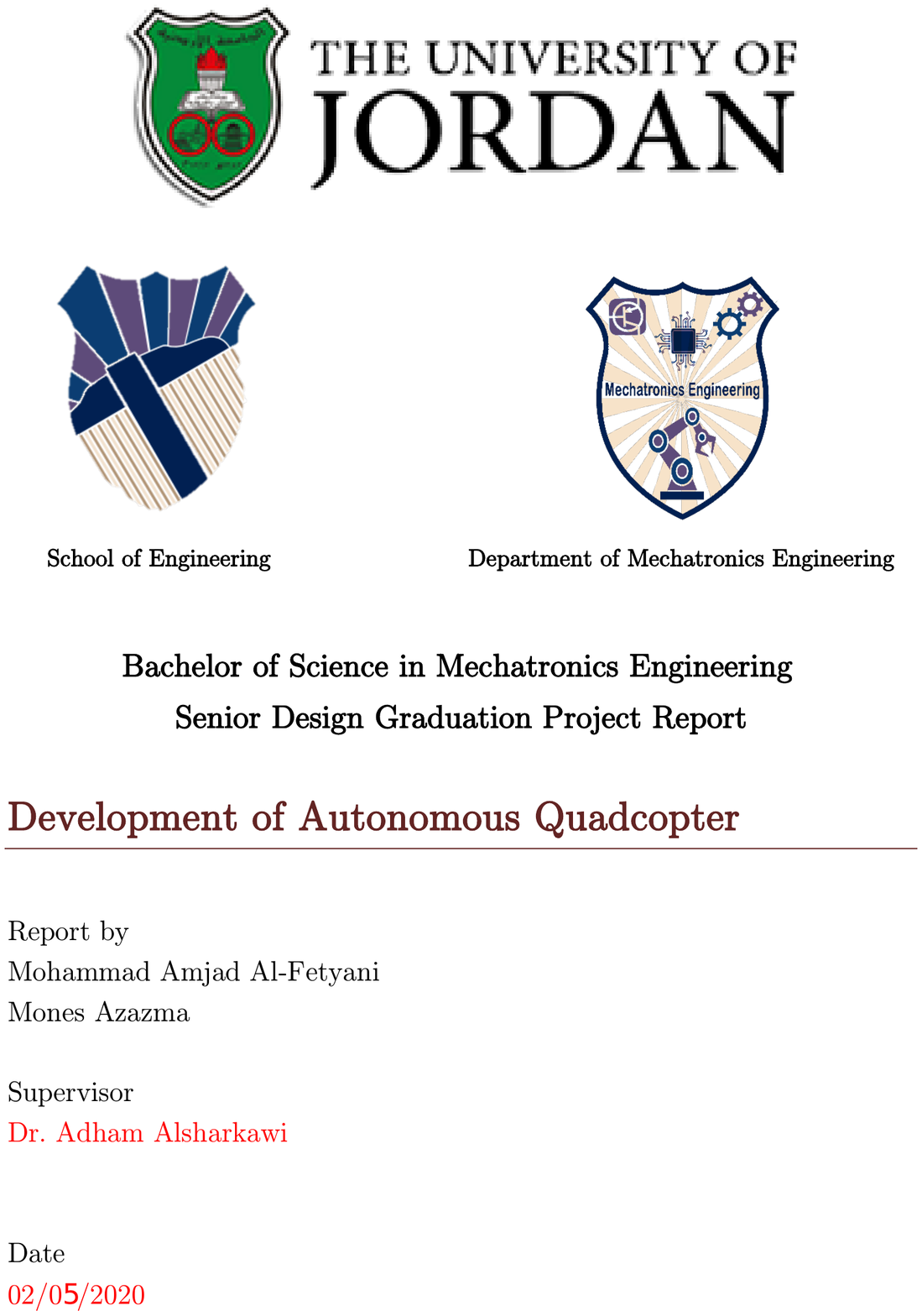}
\pagenumbering{roman}
\raggedbottom

\addtocontents{toc}{\protect{\pdfbookmark[0]{\contentsname}{toc}}}
\tableofcontents
\listoffigures
\listoftables
\chapter{Introduction}
\pagenumbering{arabic}
\section{Background}
Robotics is an interdisciplinary branch of engineering and science that deals with systems in which robots do more than what is said on papers as robotics focus on safety, flexibility, and ease of operation. A robot can be a simple box with wheels, or a complex robot consisting of a few modules or a set of modules. A robot is mainly composed of three components, sensing devices that produce signals which determine the current state of itself, the hardware and firmware which processes and manages these signals and outputs them to a hardware, and an actuator that handles the robot's capabilities and functions. This involves complex mathematical and algorithmic processes, and this has been shown to have a high value for innovation, growth, productivity, competitiveness and cost saving.\\ 

Robotics is a fast developing branch of engineering with a variety of applications that are important for the industry, academia, government, and the military. Robotics is also a fast-growing field that is changing the way people live and work, and it will change the world. The field has been in constant growth and development since the 1980s. Over the past three decades, robotics has reached its potential and become the next generation of engineering tools.\\ 

One of the major reasons for the growth of robotics in recent years is the growing interest in robotics for the commercial sector. Robots in the commercial market are gaining popularity in areas such as construction, manufacturing, medical services, and more. Robotics are increasingly becoming an important part of a company's operations and can improve quality and lower costs. In addition, robots in the commercial space are beginning to be used to assist humans in their work.\\ 

Many companies have used robots to assist workers in their operations in recent years. Companies like Amazon have used robots to pick up and deliver packages to customers. Google's self-driving cars are now being used in an extensive way in many U.S. cities. In the medical field, the use of robots for diagnostic purposes has become an increasing trend. While the technology is still in its early stages, it is possible that this will lead to significant improvements in the diagnosis and treatment of patients.\\ 

Most robots can be classified as "Autonomous Robotics", with autonomous being a relative term and more accurately referring to the robot being able to make the correct decision in response to a given situation, rather than being programmed with that decision. Most robots can be further classified as "Intelligent Machines" meaning that they can recognize their environment, recognize their surrounding and act on that information. The development of such robots has been a breakthrough in how robotics technology has progressed as a whole and can be seen as one of the first major steps towards developing a truly intelligent robot.\\ 

There are many types of robots, from simple robots like those built for domestic use to more complex robots like those used for research, construction and military applications. While they may be built of many different materials, they are usually constructed of many different components. Each robot is usually designed for a specific task and a specific environment. For example, an industrial robot, such as a Robotic Manipulator (R.M.), may be used in industrial or medical settings, while a robotic pet robot, such as a Rover, may be used for personal or companion care tasks.\\ 

Drones can be classified into three main categories according to their design; fixed-wing, single rotor and multirotor. These classifications are intended to cover the vast majority of drones. Fixed-wing drones are generally used by the military or police in support of troops on the ground. This type of aircraft is most commonly used for surveillance and patrol, although some can be used for attack or intelligence gathering missions.\\ 

Single rotor drones, also known as helicopters, generally have a long rotary blade and are used for air support and air cover over large areas. They are also used for military and paramilitary operations, reconnaissance and monitoring and are generally much more expensive than fixed-wing drones. This type of drone is commonly used by law enforcement agencies around the world. And they provide a very wide array of capabilities and are not usually flown by hobbyists. This is because they are generally large, complex, expensive to fly and have a limited flight range.  \\

For the purpose of this project, the focus will be on the multirotor drones. Multirotor drones are very popular among the drone hobbyist due to the fact that they make great aerial photography and filming equipment. They are small, light and easy to control. They are also very easy to fly, very agile and they can carry a fair amount of payload. Multirotors can be controlled via a remote controller or via an app on a smartphone. There are many types of multirotors like quad, hexa and octa rotor, but the attention needs to be focused on the one that most resemble the "normal" multirotor that is the quadrotor.\\ 

Quadrotor or quadcopter is the most common multirotor because it has four rotors that can be mounted in a quad equally spaced form, making it the most simple multirotor model. Although, quadrotors come with different sizes that are generally small, but they are very powerful and capable of providing high speed that allows it to fly very fast, easily and very well. A quadcopter can be controlled by joysticks in different ways that are convenient and it has many flight modes. A flight controller is essential to flying a quadrotor because the rotors need to be controlled.\\ 

A flight controller is the most important part of a quadrotor that makes it fly. it is usually the part that can be mounted on the quadrotor and it can be a small electronic device, that has a gyroscope and other sensors connected to it to help the quadrotor fly. It can be used to control the position of the quadrotor or it can be used to control the orientation of the quadrotor to steer it in a certain direction.\\ 

These flight controllers are extremely common on the market and allow the user to control the quadcopter by simply selecting the desired flight mode by pressing a button on the remote control. This means the user must know the desired mode to select the desired controller. The flight controller should also be equipped with a receiver that can receive the necessary commands from a remote controller and propagate them to the quadrotor to fly.\\ 

Controlling a quadcopter is not an easy task and requires a good understanding of how a quadcopter works. There are different types of functions that a quadcopter can have, like hovering or rolling, and there are different commands that can be changed on the fly. There are many different types of control systems that are used in a quadcopter and these control systems are used to control different functions of a quadcopter. This is considered a problem as the control approach and the controller parameters of a quadcopter can get really confusing since there are many types of parameters in a control system.\\ 

In the next few sections, this problem and other problems are going to be discussed in further details as well as a description of what needs to be achieved and the different controls and mechanisms that are involved.\\ 

\section{Problem definition}
This work focuses on two problems encountered in the field of drones, particularly quadcopters. The main problem is the fact that it is almost impossible to integrate advanced control technologies into ready-made autopilots. It is also hard to find a work that includes clear and straightforward steps on how to design a quadcopter controller, so this work will tackle both problems.

\section{Literature review}
In the last few years, the field of drones has received a lot of incredible attention from researchers. From autonomous drones to robotic soldiers to aerial surveillance, there has never been a better time to be a drone geek. This focus has largely been due to advancements in technologies and algorithms for aerial robotics. These include such applications as automated search and rescue and surveillance drones. It is likely, however, that we are still many years away from a day where drones are used to safely drop aid kit into a patient's hospital room or to inspect the contents of a house to detect and assess a potential threat to humans.\\ 

One of the major areas of research is how to make drones work more efficiently and make them safer to operate. Many people think of drones as the ultimate solution to the issue of search and rescue, and many of the issues are already being solved by companies that design and build drones. The technology itself is not new and it was only a few years ago that we saw the first video from Amazon of a flying drone delivering a package to the front door of a house.\\ 

In order to provide a more detailed explanation of what has been achieved so far, it is better to examine a couple of researches. First, consider one of the more widely-read research paper, by \citet*{mahony2012multirotor}, which outlines a simple approach of controlling a quadcopter. This paper is quite well-known because it is used by many hobbyists to get started with quadrotors as the authors provide a tutorial introduction to modeling, estimation, and control for a quadrotor.\\ 

A second research is by \citet*{sabatino2015quadrotor} describing the development of a more complex quadrotor control. In the paper, the authors first derive a mathematical model of a quadrotor, and then discusses how to linearize the derived mathematical model. Finally, the authors discusses how the quadrotor can be controlled using the Linear Quadratic Regulator (LQR) controller. The LQR controller is shown to be a simple, easy to construct, and robust controller that is easy to integrate and operate. Other related works to the LQR controller are discussed in \cite{argentim2013pid, kuantama2018feedback, jivrinec2011stabilization}.\\ 

Another research is by \citet*{bresciani2008modelling} describing the development of another linear controller, PID controller, that is considered as one of the most widely used controller. This is due to its simplicity, low number of parameters, and low cost. It is very simple to implement as well. The authors of this paper have successfully demonstrated that an integrated PID controller is able to perform a range of basic operations for a single-pilot quadcopter in a reasonable amount of time. The paper also discusses a range of topics regarding PID controller, including control theory, simulation, application, integration and hardware optimization. Many related studies have been performed to validate the PID controller including \cite{praveen2016modeling, leong2012low, kodgirwar2014design, paiva2016modeling}.\\ 

Another more complex linear controller is the Model Predictive Control (MPC) that is designed to be highly robust and reliable. The MPC is discussed in a number of papers, including a paper by \citet*{islam2017dynamics}, where they describe how the MPC can be used to make the quadcopter track a desired trajectory effectively, and many other papers like \cite{ganga2017mpc, bemporad2009hierarchical, iskandarani2013linear}.\\ 

Linear controllers are considered a great choice to implement a low cost and high quality quadcopter control system as they require minimal software and are relatively easy to tune. However, it is important to choose the correct region of operation since linear controllers suffer from performance degradation when operated in different regions \cite{rudol2016bridging, aangenent2008nonlinear}. This is why nonlinear controllers are considered to be a better choice for a quadcopter control.\\ 

Nonlinear controllers are considered to be more flexible and capable in application areas such as obstacle avoidance and precision navigation because of the different region of operation. Nonlinear controllers are discussed in many works in the literature, but few of these works are based on experimental design and thus it is difficult to provide a quantitative assessment of the performance and reliability of such controllers.\\ 

One of the most notable experiments on the performance and reliability of nonlinear controllers is provided by \citet*{bouabdallah2005backstepping}. The experiment was designed to evaluate the performance of a sliding mode and a backstepping controllers. The authors observed that the performance of the backstepping controller was significantly better than the sliding mode controller. These are well known types of nonlinear controllers and many studies have been performed on them, including \cite{xu2006sliding,xu2008sliding,sumantri2014second,sridhar2017non} for the sliding mode controller, and \cite{madani2006backstepping,madani2006control,fang2011adaptive,madani2007sliding} for the backstepping controller.\\ 

In an analogous experiment, \citet*{dydek2012adaptive} investigated the performance and reliability of an adaptive nonlinear controller. The controller used in their study was able to successfully maintain safe operation and smooth landing of a quadcopter. In their experiment, the stability proof of the proposed controller was based on the Lyapunov approach. Adaptive controllers have also been proposed in \cite{nicol2011robust, zhao2014nonlinear,palunko2011adaptive}.\\ 

Another nonlinear controller that is explored a lot in the literature is based on fuzzy logic which was first proposed by \citet{zadeh1965fuzzy}. One paper is by \citet*{santoso2015fuzzy} that shows the power use of a fuzzy controller to enhance the performance of a quadcopter. The authors developed a self-tuning fuzzy controller for trajectory tracking in the absence of complex mathematical formulations. Although this approach has gained popularly due to a few recent papers like \cite{olivares2012quadcopter, talha2019fuzzy, kuantama2017pid}, it is mostly known that it requires an expert to design a well performing fuzzy logic controller.\\ 

This problem of designing a fuzzy controller was addressed by \citet*{jang1993anfis} where the author presents an adaptive network based fuzzy inference  system (ANFIS) to construct input-output mapping. ANFIS can generate membership functions and refine fuzzy if-then rules without the need of an expert. Although this approach still does not receive a lot of attention as a research topic, but a few recent papers were able to use ANFIS to produce a powerful fuzzy controller as in \cite{anjum2016attitude, santoso2016adaptive, dorzhigulov2018anfis, ponce2016fuzzy}.\\ 

In a recent study by \citet*{li2017deep}, a nonlinear controller based on deep neural networks was proposed that has the potential to improve the tracking precision of a quadcopter. Their study also includes an experimental investigation on the potential of such controller to increase the tracking accuracy in a real-world quadcopter. A drawback of this approach is that the improvements are highly dependant on the training data that is provided to the deep neural networks.\\ 

The approach of controlling a quadcopter using neural networks has been further studied by
\citet*{boudjedir2012adaptive}. Where here, the goal is to enhance the accuracy of controlling a quadcopter at the presence of sinusoidal disturbance. A similar approach is proposed by \citet*{efe2011neural} to handle disturbances using neural networks that emulate the performance of a fractional order PID controller. The author concluded that this technique can be extended to emulate a large class of intelligent systems like fuzzy logic, which in turn may provide extensive flexibility for microprocessors.\\ 

More recently, a new approach has been proposed by \citet*{siti2019new} that aims to optimize the gains of a set of PD/PID controllers using genetic algorithm techniques. The authors focused their attention on the optimization problem while introducing a novel control strategy to deal with the under actuation of the quadcopter. Similar approach of using genetic algorithm has also been discussed in \cite{noshahri2014pid,nemes2015genetic,pratama2019attitude}.\\ 

To conclude, a brief review of the most controllers used to operate a quadcopter \cite{zulu2016review} reveals that each of them have different characteristics and can be used for different tasks. The optimal selection of one controller is not necessarily the optimal selection for the whole quadcopters. To find an optimal controller for a particular problem, a good knowledge of the application is required.\\ 

\section{Aims and objectives}
This project aims to introduce a simple and clear way on how to integrate advance control systems into ready-made autopilots. It also aims to provide clear instructions and steps on how to design a quadcopter controller.\\

The objectives of this work are as follows:
\begin{enumerate}
    \item Establish the mathematical model of the quadcopter.
    \item Obtain a linearized version of the mathematical model.
    \item Design simple and advanced controller for stabilizing the quadcopter.
    \item Deploy the developed controller to hardware.
\end{enumerate}

\section{Project plan}
\begin{figure}[h]
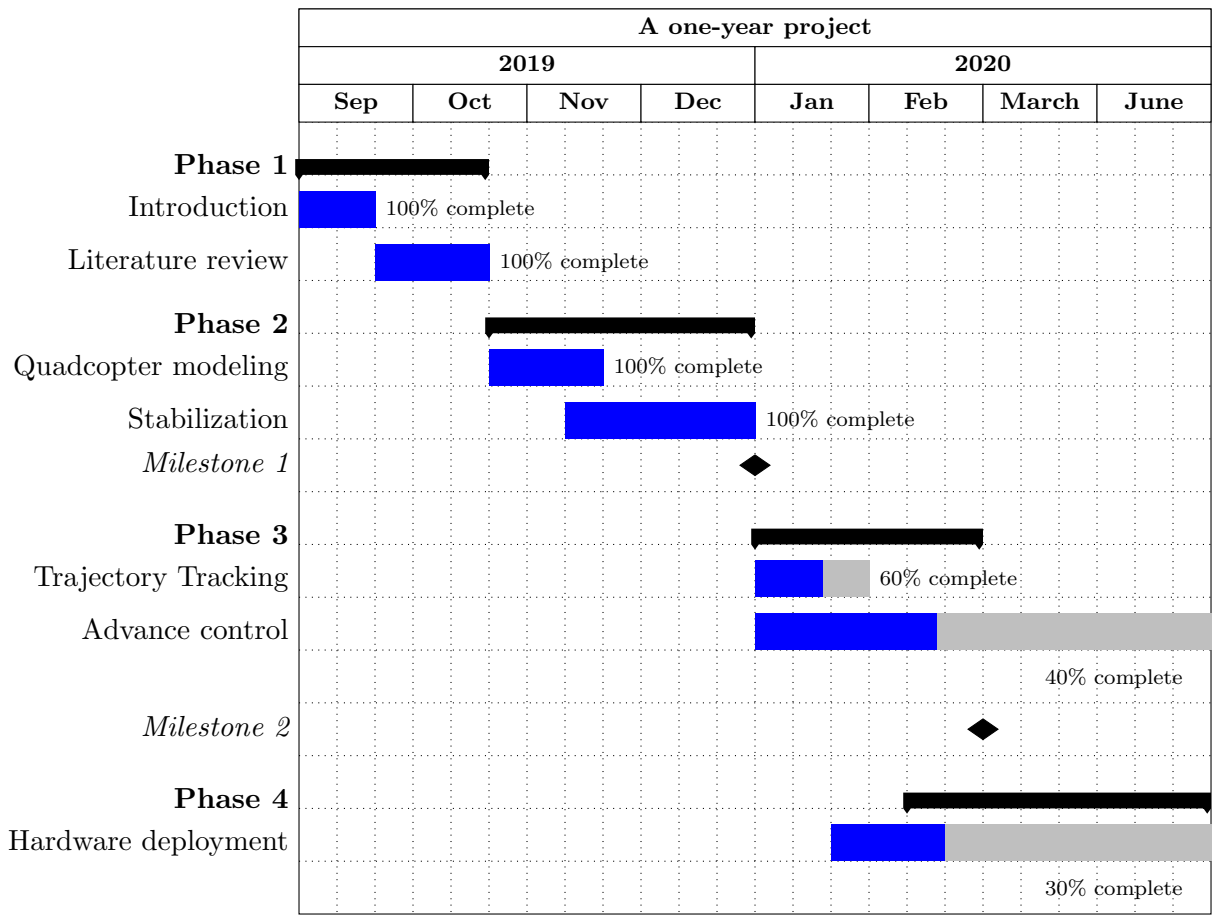

    \centering
     \begin{ganttchart}[
     y unit title=0.5cm,
     y unit chart=0.7cm,
     vgrid,hgrid,
     title height=1,
     title label font=\bfseries\footnotesize,
     bar/.style={fill=blue},
     bar height=0.7,
     group right shift=0,
     group top shift=0.7,
     group height=.3,
     group peaks width={0.2},
     inline]{1}{24}
    \gantttitle{A one-year project}{24}\\  
    \gantttitle[]{2019}{12}                 
    \gantttitle[]{2020}{12} \\              
    \gantttitle{Sep}{3}                      
    \gantttitle{Oct}{3}
    \gantttitle{Nov}{3}
    \gantttitle{Dec}{3}
    \gantttitle{Jan}{3}
    \gantttitle{Feb}{3}
    \gantttitle{March}{3} 
    \gantttitle{June}{3}\\
    \ganttgroup[inline=false]{Phase 1}{1}{5}\\ 
    \ganttbar[progress=100,inline=false]{Introduction}{1}{2}\\
    \ganttbar[progress=100,inline=false]{Literature review}{3}{5}\\

    \ganttgroup[inline=false]{Phase 2}{6}{12} \\ 
    \ganttbar[progress=100,inline=false]{Quadcopter modeling}{6}{8} \\
    \ganttbar[progress=100,inline=false]{Stabilization}{8}{12} \\
    \ganttmilestone[inline=false]{Milestone 1}{12} \\       
    
    \ganttgroup[inline=false]{Phase 3}{13}{18} \\ 
    \ganttbar[progress=60,inline=false]{Trajectory Tracking}{13}{15} \\ 
    \ganttbar[progress=40,inline=false, bar progress label node/.append style={below left= 10pt and 7pt}]{Advance control}{13}{24} \\ \\
    \ganttmilestone[inline=false]{Milestone 2}{18} \\
    \ganttgroup[inline=false]{Phase 4}{17}{24} \\ 
    \ganttbar[progress=30,inline=false,bar progress label node/.append style={below left= 10pt and 7pt}]{Hardware deployment}{15}{24}\\ 
\end{ganttchart}
    \caption{Gantt diagram for the project.}
    \label{fig:gant}
\end{figure}

The plan of this project is presented as Gantt diagram in \figurename{ \ref{fig:gant}}.

\section{Report organization}
The rest of this work is organized as follows: Chapter \ref{ch1} ( \nameref{ch1}), Chapter \ref{ch2} (\nameref{ch2}), Chapter \ref{ch3} (\nameref{ch3}), Chapter \ref{ch3} (\nameref{ch3}), Chapter \ref{ch4} (\nameref{ch4}), Chapter \ref{ch6} (\nameref{ch6}), Chapter \ref{ch7} (\nameref{ch7}) and Chapter \ref{ch8} (\nameref{ch8}).

\chapter{Design considerations}
\label{ch1}

This chapter discusses the available design options and justifies the approach adopted in this work. It also presents the constraints and the standards that are encountered throughout the work.

\section{Design options}
Many techniques have been developed to model physical systems including mathematical modeling, data modeling and agent-based model. Mathematical modeling is normally based on system dynamic equations, data modeling is based on system identification and agent-based model normally involve optimization problems. \\

In this work, mathematical modeling is adopted because it is capable to accurately predict the behaviour of the quadcopter. In addition, it facilitates the design of the controllers based widely known control techniques. \\

Many types of control techniques are used nowadays to control the quadcopter such as PID, LQR, SMC and many more. However, this work mainly focuses on two types of controllers, PID and Fuzzy. Fuzzy controller is a nonlinear, easy to construct and robust controller, which is why it is adopted in this work. Because of the simplicity of the PID controller, it is used as a baseline to compare the performance of the fuzzy controller. \\

Although, there are many ways to implement the mathematical model in software such as programming the model using different programming language (e.g Python and C++) but the one that is most flexible and reliable is the MATLAB software, which is adopted in this work. One advantage of MATLAB software is that it does not require you to discretize the mathematical model like in programming. It also contains an incredible tools that would ease the controller design process.

\section{Design constraints and standards}
Throughout this work, MATLAB software version 2019a is used for simulation, control design and deployment. The international standard that MATLAB software uses for Simulink is "IEC 61508" and for the model-based designer it uses the standard "IEC 61508" \cite{matlab}. \\

One of the most important constraints in this work is the control signal. The control signal has to be physically possible and because it represents a PWM signal, it should be between 1000 and 2000. Another constrain is the budget, the design of the controller must be compatible with the design of a simple quadcopter.
\chapter{Quadcopter modelling}
\label{ch2}
This chapter presents a mathematical model for a three-dimensional quadcopter. A mathematical model can help us understand how a complex system behaves and what the consequences of that behavior are. The derivation of the mathematical model is achieved in a step-wise manner throughout this work as we proceed from the simplest model to the most complex. As the model becomes more complex, it allows us to predict the behavior of the realistic quadcopter with much higher confidence.\\ 

The first section of this chapter (\ref{sec:s11} \nameref{sec:s11}) describes the most common movements of a quadcopter, shows the main idea behind each movement and how it can be achieved, and gives a visual explanation of the quadcopter mechanisms. This section also describes different configurations of the quadcopter.\\ 

The second section of this chapter (\ref{sec:s12} \nameref{sec:s12}) shows the math behind the quadcopter movements, explains what are the inputs and outputs of such system, and gives a ready-to-use model that can be easily simulated and modified to meet the needs of different projects.\\ 


\section{Flight mechanism}
\label{sec:s11}
A quadrotor is a multi-rotor aircraft with four rotors attached to its main body in an equally spaced formation. It is a lightweight flying machine that can maneuver in any direction and is capable of high speeds. It is also capable of vertical take-off/landing (VTOL) and can be used to carry a heavy load of about 1 kg. A significant advantage of this vehicle is its versatility; it is able to operate in a variety of different configurations, making it easily adaptable to a wide range of missions.\\ 

\begin{figure}[h]
    \centering
    \includegraphics[width=0.9\textwidth]{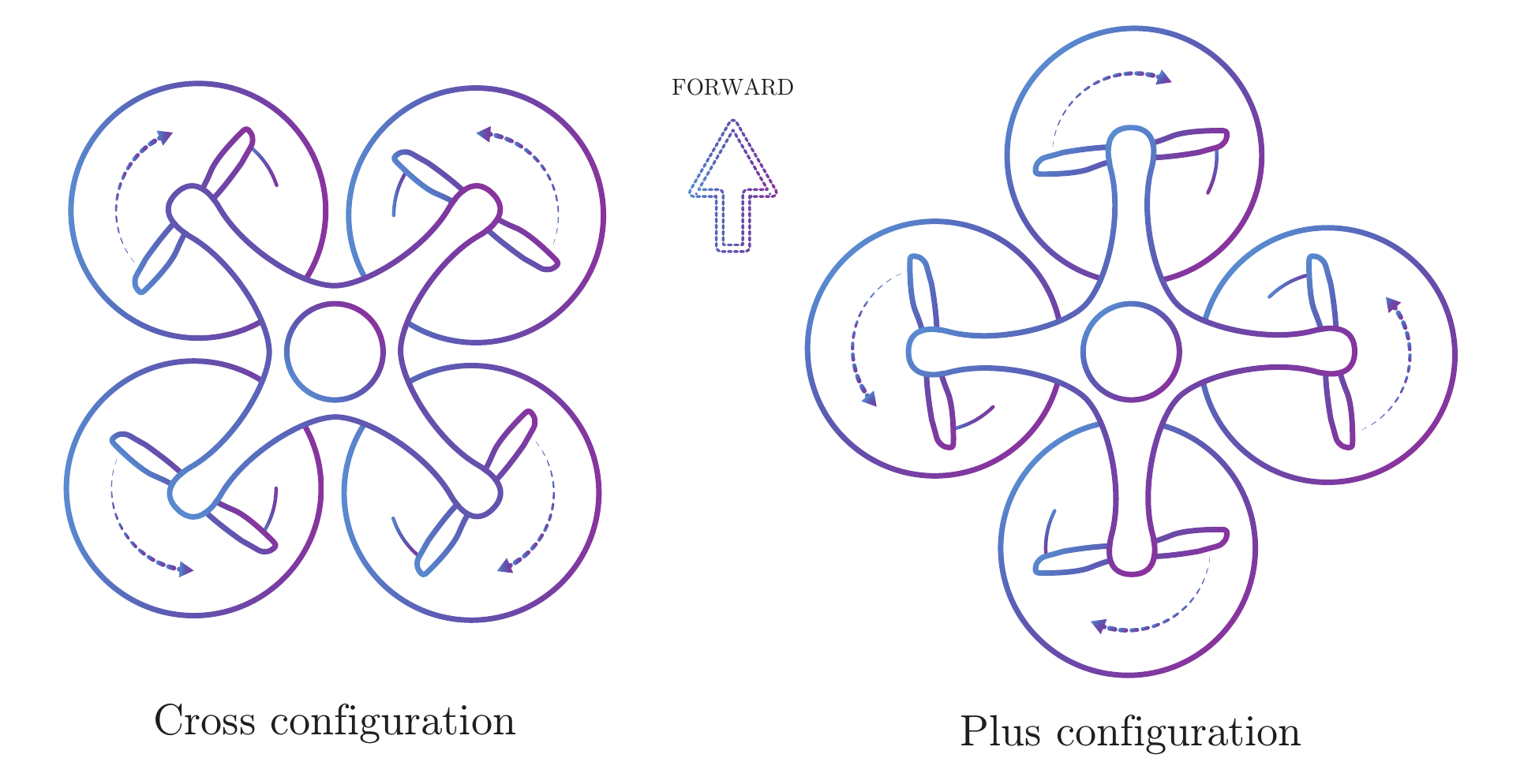}
    \caption{Quadcopter configurations.}
    \label{fig:config}
\end{figure}

A quadcopter is mainly composed of four movements, throttle, roll, pitch, and yaw. These movements are achieved in different manners depending on the configuration selected. There are many configurations of a quadcopter but the two most common configurations are a cross 'X' configuration, and a plus '+' configuration. Both configurations are shown in \figurename{ \ref{fig:config}}, where the cross configuration is shown on the left and plus configuration is shown on the right.\\ 

In general, the cross configuration is more stable than the plus configuration \cite{thu2017designing} due to the fact that all four motors contribute to the stability of the quadrotor. The cross configuration is also considered the most popular configuration \cite{agrawal2015multi} since a camera can point out between the front two motors which makes it helpful to be used for aerial photography/videography. The plus configuration, on the other hand, is easier to control compared to the cross configuration \cite{agrawal2015multi} because only two motors control roll and pitch movements which makes it a great choice for acrobatic flying \cite{niemiec2016comparison}.\\ 

\begin{figure}[ht]
    \centering
    \includegraphics[width=0.4\textwidth]{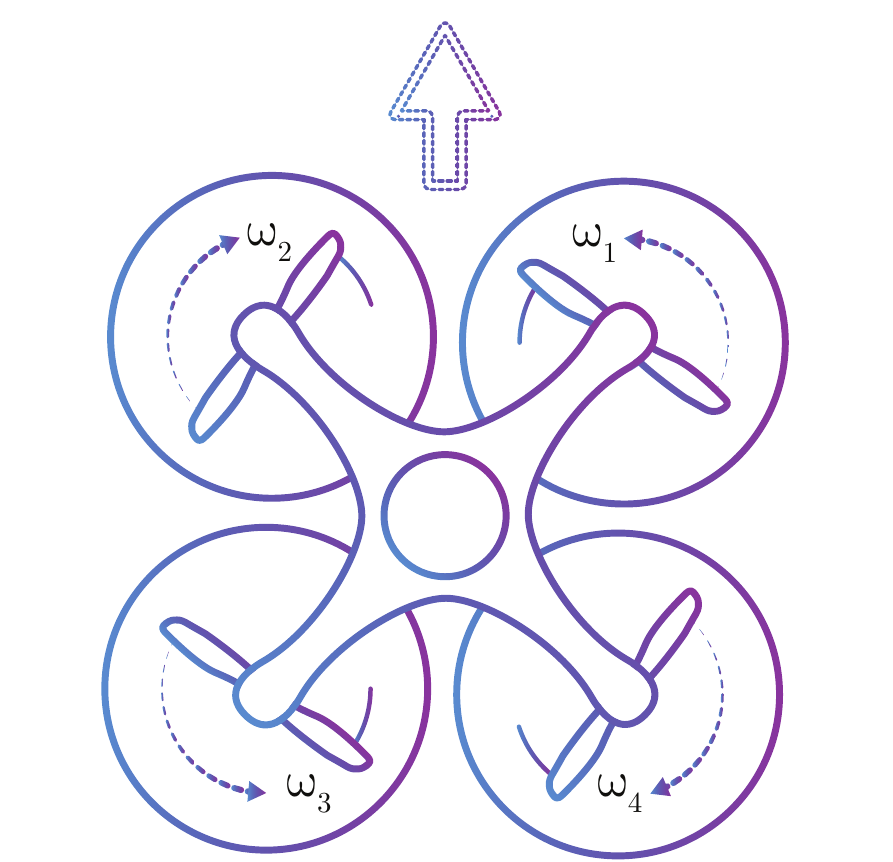}
    \caption{Quadcopter cross configuration.}
    \label{fig:config1}
\end{figure}

This work focuses on the cross configuration and thus all four movements are going to be explained based on the cross configuration. These four movements are described in terms of motors angular velocity, $\omega$, that are labeled from 1 to 4 as presented in \figurename{ \ref{fig:config1}}.\\ 

The throttle movement is used to control the height, up or down, of the quadrotor. It is achieved by changing all angular velocities of the motors, increasing or decreasing. The amount of thrust applied to the motors is directly proportional to the height of the quadcopter. For instance, to keep the quadrotor hovering at a desired height, an equal amount of thrust should be applied to the motors.\\ 

The roll movement is a rotation around the front-to-back axis of the quadrotor and is achieved by increasing or decreasing the angular velocities of motor 2 and motor 3 ($\omega_2$ and $\omega_3$) and at the same time decreasing or increasing the angular velocities of motor 1 and motor 4 ($\omega_1$ and $\omega_4$). This movement is used to move the quadrotor to the right or to the left.\\ 

The pitch movement is a rotation around the side-to-side axis of the quadrotor and it is very similar to the roll movement. The only difference is that it is used to move the quadrotor forward or backward. And thus, it is achieved by increasing or decreasing the angular velocities of motor 3 and motor 4 ($\omega_3$ and $\omega_4$) and at the same time decreasing or increasing the angular velocities of motor 1 and motor 2 ($\omega_1$ and $\omega_2$).\\ 

It is to be noted that the roll and pitch movements stand in the way of the throttle movement. This due to the fact that the motors are no longer on a horizontal plane and thus they are not generating an upward force. To counter this behavior, the angle of rotation must be taken into consideration in the throttle movement.\\ 

The last and final movement is the yaw movement which is a rotation around the vertical axis of the quadcopter. The yaw movement is acquired by increasing or decreasing the angular velocities of two opposite motors (say $\omega_1$ and $\omega_3$ and at the same time decreasing or increasing the angular velocities of other two (say $\omega_2$ and $\omega_4$).\\ 

\section{Basic concepts}
\label{sec:s12}
\begin{figure}[ht]
    \centering
    \includegraphics[width=\textwidth]{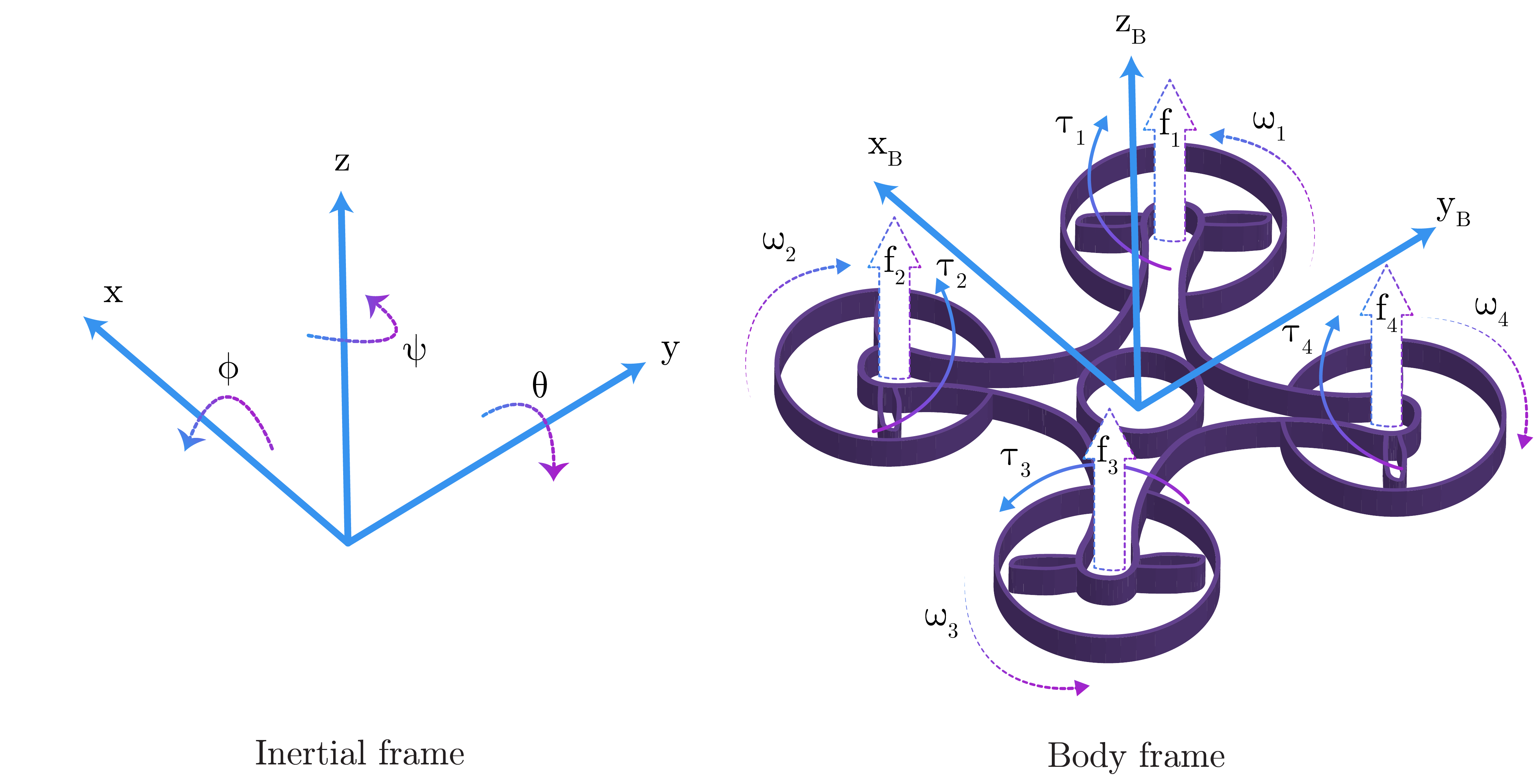}
    \caption{Quadcopter reference frames.}
    \label{fig:config2}
\end{figure}

Before describing the dynamical model of a quadcopter, it is necessary to introduce two reference frames. The first frame is kept fixed and is referred to as a world or inertial frame. This frame is used to track the movements of the second frame. The second frame is referred to as body frame and is associated with the quadcopter, hence it moves with the quadcopter. These two frames are assumed as presented in \figurename{ \ref{fig:config2}}.\\ 

The absolute position of the quadcopter is introduced in the world frame as $x$, $y$ and $z$ axes and these are contained in a column vector, $\bm{\xi}$. The attitude, or the angular position, of the quadcopter is defined in the world frame as $\bm{\Theta}$, where the roll angle is $\phi$, the pitch angle is $\theta$, and the yaw angle is $\psi$. These three angles are referred to as Euler angles.\\ 
\begin{equation}
    \bm{\xi} = 
    \begin{bnmatrix}
         x \\ y \\ z
    \end{bnmatrix}
    , \quad
    \bm{\Theta} = 
    \begin{bnmatrix}
         \phi \\ \theta \\ \psi
    \end{bnmatrix}
\end{equation}

The body frame is located at the center of mass of the quadcopter. The linear velocities for the three axes are defined in the body frame by $\bm{V_B}$ and the corresponding angular velocities are denoted by $\bm{v}$. The angular velocities in the $x$, $y$ and $z$ axes are denoted by $p$, $q$ and $r$ respectively.\\ 
\begin{equation}
    \bm{V_B} = 
    \begin{bnmatrix}
         V_{B,x} \\ V_{B,y} \\ V_{B,z}
    \end{bnmatrix}
    , \quad
    \bm{v} = 
    \begin{bnmatrix}
         p \\ q \\ r
    \end{bnmatrix}
\end{equation}

To keep track of the state of the quadcopter, the positions and the angular velocities of the quadcopter that are measured in the body frame should be expressed in the world frame. For the positions, this mapping between the two frames can be achieved through a rotation matrix of each axis. These rotation matrices are as follows:

\begin{equation}
    \bm{R_{z,\psi}} = 
    \begin{bnmatrix}
    \text{cos}(\psi) & -\text{sin}(\psi) & 0 \\
    \text{sin}(\psi) & \text{cos}(\psi) & 0\\
    0 & 0 & 1
    \end{bnmatrix}
\end{equation}
\begin{equation}
    \bm{R_{y,\theta}} = 
    \begin{bnmatrix}
    \text{cos}(\theta) & 0 & \text{sin}(\theta) \\
     0& 1 & 0\\
    -\text{sin}(\theta) & 0 & \text{cos}(\theta)
    \end{bnmatrix}
\end{equation}
\begin{equation}
    \bm{R_{x,\phi}} = 
    \begin{bnmatrix}
    0 & 0 & 1\\
     0 & \text{cos}(\phi) & -\text{sin}(\phi)\\
    0 & \text{sin}(\phi) & \text{cos}(\phi) 
    \end{bnmatrix}
\end{equation}
where $\bm{R_{k,t}}$ is the rotation matrix about axis $k$ with angle $t$.\\ 

The over all rotation matrix from the body frame to the world frame is derived from this as follows:
\begin{equation}
    \bm{R} =\bm{R_{z,\psi}R_{y,\theta}R_{x,\phi}}
\end{equation}

\begin{equation}
    \bm{R} = 
    \begin{bnmatrix}
        C_{\psi} C_{\theta} & {C_{\psi} S_{\theta} S_{\phi}-S_{\psi} C_{\phi}} & {C_{\psi} S_{\theta} C_{\phi}+S_{\psi} S_{\phi}} \\[2pt]
        {S_{\psi} C_{\theta}} & {S_{\psi} S_{\theta} S_{\phi}+C_{\psi} C_{\phi}} & {S_{\psi} S_{\theta} C_{\phi}-C_{\psi} S_{\phi}} \\[2pt]
        {-S_{\theta}} & {C_{\theta} S_{\phi}} & {C_{\theta} C_{\phi}}
    \end{bnmatrix}
\end{equation}
where $C_t = \text{cos}(t)$ and $S_t = \text{sin}(t)$.\\ 

The inverse of the rotation matrix $\bm{R}$ is the rotation matrix from the world frame to the body frame. It can be calculated from $\bm{R^{-1} = R^T}$ due to the orthogonality of the rotation matrix $\bm{R}$.\\ 

The angular velocities of the quadcopter, which are measured in the body frame, are not the exact same values as seen from the world frame. They should be mapped to the world frame throughout the rotation matrix $\bm{\eta}$ \cite{bresciani2008modelling}.
\begin{equation}\label{eq3}
    \bm{\dot{\bm{\Theta}}} = \bm{\eta v}, \quad
    \begin{bnmatrix}
    {\dot{\phi}} \\ {\dot{\theta}} \\ {\dot{\psi}} 
    \end{bnmatrix}
    =
    \begin{bnmatrix}
       {1} & {S_{\phi} T_{\theta}} & {C_{\phi} T_{\theta}} \\ {0} & {C_{\phi}} & {-S_{\phi}} \\ {0} & {S_{\phi} / C_{\theta}} & {C_{\phi} / C_{\theta}}
    \end{bnmatrix}
    \begin{bnmatrix}
       {p} \\ {q} \\ {r}
    \end{bnmatrix}
\end{equation}
where $T_t = \text{tan}(t)$.\\

The motors generate a vertical force, $F$, in the body z-axis while rotating. This force is proportional to the square of a motor's angular velocity. All motors also contribute in generating a torque, $\bm{\tau}$, on all three axes in the body frame while rotating. This torque is also proportional to the angular velocities of the four motors.
\begin{equation}\label{eq:force}
F = \sum_{i=1}^{4}f_i = C_T\sum_{i=1}^{4}\omega^2_i, \quad \bm{F_B} = 
\begin{bnmatrix}
0 \\ 0 \\ F
\end{bnmatrix}
\end{equation}
where $f_i$ is the force generated by the $i^{th}$ motor, $C_T$ is the thrust coefficient and $\bm{F_B}$ is the vector representation of these forces.

The torques that are created by the four motors consist of $\tau_\phi$, $\tau_\theta$ and $\tau_\psi$ in the direction of the body $x$, $y$ and $z$ axes.
\begin{equation}\label{eq:torques}
\bm{\tau} =
\begin{bnmatrix}
\tau_\phi \\ \tau_\theta \\ \tau_\psi 
\end{bnmatrix}
=
\begin{bnmatrix}
dC_T(\omega_1^2-\omega_2^2-\omega_3^2+\omega_4^2) \\[2pt]
dC_T(\omega_1^2+\omega_2^2-\omega_3^2-\omega_4^2) \\[2pt]
C_D(\omega_1^2-\omega_2^2+\omega_3^2-\omega_4^2)
\end{bnmatrix}
\end{equation}
where $d$ is the arm length of the quadcopter and $C_D$ is the aerodynamic drag coefficient.

\section{Dynamical model}
The dynamical model of a quadrotor can be described as a set of six equations, in which three equations describe the $x,y,z$ position of the quadrotor and the other three describe the roll,pitch,yaw rotation of the quadcopter. Each of these equations takes the angular velocities of the four motors, $\omega$, as inputs and outputs the next state of the quadcopter.\\ 

In this work, several assumptions are made to simplify the derivation of the dynamical model. These assumptions do not perfectly reflect the real quadcopter but they are considered good approximations. These assumptions are as follows:

\begin{enumerate}
    \item The mass of the quadcopter is evenly distributed and is kept constant, hence its derivative is zero.
    \item The quadcopter is a rigid body, meaning that it does not deform.
    \item The center of gravity is located in the middle of the quadcopter and coincides with the origin of the body frame.
    \item The quadcopter structure is symmetrical with four arms aligned with the body x- and y- axes.
\end{enumerate}{}

Because of the assumptions we made, the dynamics of the quadcopter can be described by Newton-Euler equations. Newton's equations describe the translational motion of the quadcopter and can be formulated in the inertial frame as follows:

\begin{equation} \label{eq1}
    \begin{gathered}
    m\bm{\ddot{\xi}} = \bm{G}+\bm{RF_B}, \\[10pt]
    m
    \begin{bnmatrix}
    \ddot{x} \\ \ddot{y} \\\ddot{z}
    \end{bnmatrix}{}
    =
    \begin{bnmatrix}
    0 \\ 0 \\ -mg
    \end{bnmatrix}{}
    + \bm{R}
    \begin{bnmatrix}
    0 \\ 0 \\ F
    \end{bnmatrix}
    \end{gathered}
\end{equation}
where $\bm{G}$ is the gravitational force vector.\\

It is possible to simplify (\ref{eq1}) and isolate the vector $\bm{\ddot{\xi}}$:
\begin{equation}\label{eq:ddxi}
    \begin{bnmatrix}
    \ddot{x} \\ \ddot{y} \\\ddot{z}
    \end{bnmatrix}{}
    =
    \begin{bnmatrix}
    0 \\ 0 \\ -g
    \end{bnmatrix}{}
    +
    \frac{F}{m}
    \begin{bnmatrix}
    {C_{\psi} S_{\theta} C_{\phi}+S_{\psi} S_{\phi}} \\[2pt]
        {S_{\psi} S_{\theta} C_{\phi}-C_{\psi} S_{\phi}} \\[2pt] {C_{\theta} C_{\phi}}
    \end{bnmatrix}
\end{equation}{}

The rotational motion of the quadcopter is described by Euler's equation of motion in (\ref{eq2}) and is best expressed in the body frame since control inputs are given in the body frame as well as the measurements, which are taken from on-board sensors, are easily converted to body frame. 
\begin{equation} \label{eq2}
    \begin{gathered}{}
        \bm{\tau} = \bm{J\dot{v}}+\bm{v\times Jv}, \\[10pt]
        \begin{bnmatrix}
        \tau_\phi \\ \tau_\theta \\ \tau_\psi 
        \end{bnmatrix}
        =
        \begin{bnmatrix}
        I_{xx} & 0 & 0 \\
        0 & I_{yy} & 0 \\
        0 & 0 & I_{zz}
        \end{bnmatrix}
        \begin{bnmatrix}
            \dot{p} \\ \dot{q} \\ \dot{r}
        \end{bnmatrix}
        +
        \begin{bnmatrix}
             p \\ q \\ r
        \end{bnmatrix} \times
        \begin{bnmatrix}
        I_{xx} & 0 & 0 \\
        0 & I_{yy} & 0 \\
        0 & 0 & I_{zz}
        \end{bnmatrix}
        \begin{bnmatrix}
             p \\ q \\ r
        \end{bnmatrix}, \\[10pt]
        \begin{bnmatrix}
        \tau_\phi \\ \tau_\theta \\ \tau_\psi 
        \end{bnmatrix}
        =
        \begin{bnmatrix}
            I_{xx}\dot{p} \\ I_{yy}\dot{q} \\ I_{zz}\dot{r}
        \end{bnmatrix}
        + 
        \begin{bnmatrix}
             p \\ q \\ r
        \end{bnmatrix} \times
        \begin{bnmatrix}
             I_{xx}p \\ I_{yy}q \\ I_{zz}r
        \end{bnmatrix},\\[10pt]
        \begin{bnmatrix}
        \tau_\phi \\ \tau_\theta \\ \tau_\psi 
        \end{bnmatrix}
        =
        \begin{bnmatrix}
            I_{xx}\dot{p} \\ I_{yy}\dot{q} \\ I_{zz}\dot{r}
        \end{bnmatrix}
        + 
        \begin{bnmatrix}
             I_{zz}qr-I_{yy}qr \\ 
             -I_{zz}pr+I_{xx}pr \\ 
             I_{yy}pq-I_{xx}pq \\ 
        \end{bnmatrix} 
    \end{gathered}
\end{equation}{}
where $\bm{J}$ is the inertia matrix of the quadcopter that contains the mass moment of inertia of each axis $I_{xx}, I_{yy}$ and $I_{zz}$. \\

It is possible to simplify (\ref{eq2}) and isolate the vector $\bm{\dot{v}} = [\; \dot{p},\;\dot{q},\;\dot{r}\;]^T$:
\begin{equation}\label{eq4}
    \begin{bnmatrix}
        \dot{p} \\ \dot{q} \\ \dot{r}
    \end{bnmatrix}
    =
    \begin{bnmatrix}
    \tau_\phi/I_{xx} \\ \tau_\theta/I_{yy} \\ \tau_\psi /I_{zz}
    \end{bnmatrix}
    +
    \begin{bnmatrix}
    (I_{yy}-I_{zz})qr/I_{xx} \\
    (I_{zz}-I_{xx})pr/I_{yy} \\
    (I_{xx}-I_{yy})pq/I_{zz} \\
    \end{bnmatrix}{}
\end{equation}{}
Equation (\ref{eq4}) can be transformed to the inertial frame after integration by (\ref{eq3}).\\

These Newton-Euler equations are considered simple because no external forces or torques are included and they can be written in equation format as follows:
\begin{align}
    \ddot{x} &= \frac{F}{m}(\text{cos}\psi\;\text{sin}\theta \;\text{cos} \phi +\text{sin}\psi \;\text{sin} \phi ) \\[10pt]
    \ddot{y} &= \frac{F}{m}(\text{sin} \psi \;\text{sin} \theta\;\text{cos} \phi -\text{cos} \psi\;\text{sin} \phi )\\[10pt]
    \ddot{z} &= \frac{F}{m}(\text{cos} \theta \;\text{cos} \phi )-g \label{eq:zdd}\\[10pt]
    \dot{p} &= \frac{(I_{yy}-I_{zz})qr}{I_{xx}}+\frac{\tau_\phi}{I_{xx}}\\[10pt]
    \dot{q} &= \frac{(I_{zz}-I_{xx})pr}{I_{yy}}+\frac{\tau_\theta}{I_{yy}}\\[10pt]
    \dot{r} &= \frac{(I_{xx}-I_{yy})pq}{I_{zz}}+\frac{\tau_\psi}{I_{zz}}
\end{align}{}

To summarize all equations in one page, the following are the transformation between the angler velocities in the body frame and the inertial frame written in equation format:
\begin{align}
    \dot{\phi} &= p + \text{sin}\phi\;  \text{tan}\theta \; q + \text{cos}\phi \; \text{tan}\theta\;  r \\[10pt]
    \dot{\theta} &= \text{cos}\phi \; q - \text{sin}\phi \; r\\[10pt]
    \dot{\psi} &= \text{sin}\phi/\text{cos}\theta\;  q + \text{cos}\phi/\text{cos}\theta \; r
\end{align}

\chapter{Open-loop behaviour}
\label{ch3}
This chapter presents the open-loop behaviour of the quadcopter, where various angular velocities are applied to the quadcopter. This gives an idea of how quadcopters behave in response to the applied forces and also provides an indication of how different combinations of angular velocities can yield different quadcopter movements. \\

The first section of this chapter (\ref{sec:s31} \nameref{sec:s31}) provides an overview of the architecture that is used to build the dynamics of the quadcopter and also explains each block of the architecture in more detail. \\

The second section of this chapter (\ref{sec:s32} \nameref{sec:s32}) provides a detailed description of the various angular velocities that are applied to the quadcopter to control its movements and also illustrates the state of the quadcopter in response to the applied forces through a simulation of the flight dynamics.

\tikzstyle{input} = [coordinate]
\tikzstyle{output} = [coordinate]
\tikzstyle{pinstyle} = [pin edge={to-,thin,black}]
\section{Open-loop architecture}
\label{sec:s31}
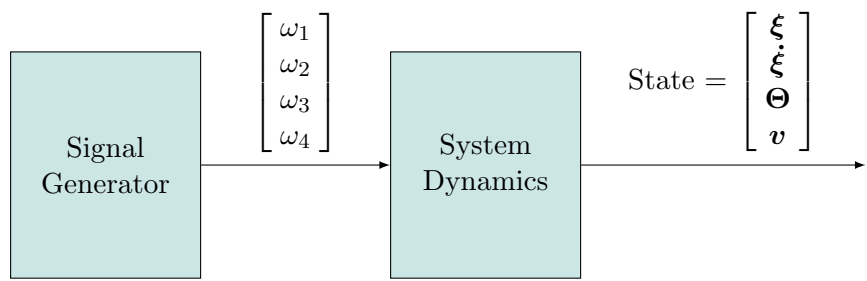
\begin{figure}[h]
    \centering
    \begin{tikzpicture}[auto, node distance=5cm,>=latex,every text node part/.style={align=center}]
    \tikzstyle{block} = [draw, fill=teal!20, rectangle, 
    minimum height=3cm, minimum width=2.5cm,node distance=5cm]
    \node [block] (controller) {Signal \\ Generator};
    \node [block, right of=controller] (system) {System \\ Dynamics};
    \draw [->] (controller) -- node {$\begin{bnmatrix}\omega_1\\\omega_2\\\omega_3\\\omega_4\end{bnmatrix}$} (system);
    \node [output, right of=system] (output) {};
    \draw [->] (system) -- node [name=y] {State = $\begin{bnmatrix}\bm{\xi}\\\bm{\dot{\xi}}\\\bm{\Theta}\\\bm{v}\end{bnmatrix}$}(output);
\end{tikzpicture}
    \caption{Open-loop architecture.}
    \label{fig:arch_open}
\end{figure}

\figurename{ \ref{fig:arch_open}} presents the open-loop architecture of the quadcopter. it contains two main blocks, one generates the angular velocities of the motors and the other gives the state of the quadcopter in response to the given signal. 

{}

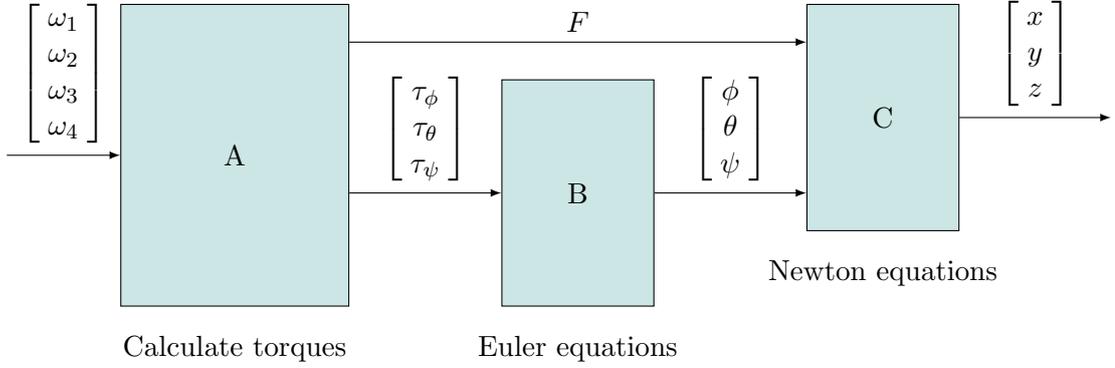
\begin{figure}[h]
    \centering
    \begin{tikzpicture}[auto, node distance=3cm,>=latex,every text node part/.style={align=center}]
    \tikzset{pinstyle/.style={pin edge={to-,thick,black}}}
    \tikzstyle{block} = [draw, fill=teal!20, rectangle, 
    minimum height=3cm, minimum width=2.5cm]
    
    \node [input, name=input] (input) {};
    \node [block,right of=input,minimum width=3cm, minimum height=4cm] (a) {A};
    \node [block, right = 2 cm of a, yshift=-0.5cm,minimum width=2cm, minimum height=3cm](b) {B};
    \node [block, right = 2 cm of b, yshift=1cm,minimum width=2cm, minimum height=3cm](c) {C};

    \node [below =0.25cm of a] (t1) {Calculate torques};
    \node [below =0.25cm of b] (t2) {Euler equations};
    \node [below =0.25cm of c] (t3) {Newton equations};
    \node [output, right of=c] (output) {};

    \draw [draw,->] (input) -- node {$\begin{bnmatrix}\omega_1\\\omega_2\\\omega_3\\\omega_4\end{bnmatrix}$} (a);

    \draw [->] ([yshift=-0.5cm]a.east) -- node  {$\begin{bnmatrix}\tau_\phi \\ \tau_\theta \\ \tau_\psi\end{bnmatrix}$}(b);

    \draw [->] ([yshift=1.5cm]a.east) -- node  {$F$}([yshift=1cm]c.west);
    \draw [->] (b.east) -- node  {$\begin{bnmatrix}\phi \\ \theta \\ \psi\end{bnmatrix}$}([yshift=-1cm]c.west);
    \draw [->] (c) -- node { $\begin{bnmatrix}x\\y\\z\end{bnmatrix}$}(output);
\end{tikzpicture}
    \caption{System dynamics architecture.}
    \label{fig:arch_dyn}
\end{figure}

The signal generator block is going to be discussed in the next section. Here, the second block, system dynamics, is discussed in further details. \figurename{ \ref{fig:arch_dyn}} illustrates the blocks arrangement inside the system dynamics main block.

{}

As noticed, it composed of three blocks. Block "A" calculates the torques from equation (\ref{eq:torques}) and directs them to block "B". It also calculates the force generated by the inputs angular velocities using equation (\ref{eq:force}) and directs them to block "C". \\

Block "B" starts by integrating the angular velocities, $\bm{v}$, after calculating the angular accelerations, $\bm{\dot{v}}$, from equation (\ref{eq4}). This is followed by transforming the angular velocities, $\bm{v}$, from the body frame to the angular velocities, $\bm{\dot{\Theta}}$, in the world frame using the transformation in (\ref{eq3}) and then integrating again to get Euler angles, $\bm{\Theta}$. \\

Finally, block "C" is a straightforward block, where it calculates the translational acceleration, $\bm{\ddot{\xi}}$, from equation (\ref{eq:ddxi}) and then double integrating to get the translational position $\bm{\xi}$.

\section{Open-loop simulation}
\label{sec:s32}
The previous architecture is implemented in MATLAB Simulink 2019a for simulation. The parameters of the quadcopter are determined experimentally, as presented in Appendix \ref{app1}, and are shown in Table \ref{tab:param}.\\

The simulation progresses at a sample time of 0.01 second to a total of 2 seconds. The state of the quadcopter, positions and velocities, are initially set to zero which means that the body frame coincides with the inertial frame and the quadcopter is initially stable.

\begin{table}[H]
    \centering
    \caption{Parameter values for simulation.}
    \label{tab:param}
    \setlength{\extrarowheight}{.95ex}
    
    \begin{tabular}{|>{\columncolor{Gray}}c|l|c|}
        \toprule
        \rowcolor{Gray}
         \textbf{Parameter} & \multicolumn{1}{c|}{\textbf{Description} }& \textbf{Value}   \\
         \midrule 
         m & Total mass of the quadcopter & 0.9 [kg] \\
         g & Gravity acceleration & 9.81 [m/s$^2$] \\
         l & Arm length & 0.21 [m] \\
         I$_{xx}$ & Moment of inertia around x axis & $1.467 \times 10^{-2}$ [kg$\times$m$^2$]\\
         I$_{yy}$ & Moment of inertia around y axis & $1.667 \times 10^{-2}$ [kg$\times$m$^2$]\\
         I$_{zz}$ & Moment of inertia around z axis & $1.325 \times 10^{-2}$ [kg $\times$m$^2$]\\
         C$_T$ & Thrust coefficient & $4.980 \times 10^{-8}$ [N/rpm$^2$]\\
         C$_D$ & Torque coefficient & $5.804 \times 10^{-9}$ [Nm/rpm$^2$]\\[2pt]
         \bottomrule
    \end{tabular}
\end{table}{}

As discussed in section \ref{sec:s11}, four main movements of the quadcopter can be achieved through different control inputs, which are the angular velocities of the four motors. These movements are achieved here in four different stages, each stage takes 2 seconds to complete. The control inputs used in each stage are as follows:
\begin{align}
    \omega_1 &=  
    \begin{cases} 
      2000sin(\pi t) + \omega_e & 0\leq t\leq 2 \\
      500sin(\pi t) + \omega_e & 2\leq t\leq 8 \\
   \end{cases}\\[10pt]
    \omega_2 &=  
    \begin{cases} 
      2000sin(\pi t) + \omega_e & 0\leq t\leq 2 \\
      250sin(\pi t) + \omega_e & 2\leq t\leq 4\; \text{or} \; 6\leq t\leq 8\\
      500sin(\pi t) + \omega_e & 4\leq t\leq 6 \\
   \end{cases}\\[10pt]
    \omega_3 &=  
    \begin{cases} 
      2000sin(\pi t) + \omega_e & 0\leq t\leq 2 \\
      250sin(\pi t) + \omega_e & 2\leq t\leq 6\\
      500sin(\pi t) + \omega_e & 6\leq t\leq 8 \\
   \end{cases}\\[10pt]
    \omega_4 &=  
    \begin{cases} 
      2000sin(\pi t) + \omega_e & 0\leq t\leq 2 \\
      500sin(\pi t) + \omega_e & 2\leq t\leq 4 \\
      250sin(\pi t) + \omega_e & 4\leq t\leq 8\\
   \end{cases}
\end{align} \\
where $\omega_e$ is the angular velocity of each motor that maintains an equilibrium state of the quadcopter which is calculated in equation (\ref{eq:we}).\\

It is to be noted that the selection of such control inputs is not arbitrary, as the angular velocities of the motors are almost smooth and because the quadcopter is unstable, the time for each stage has to be tight so that the response can be easily visualized. Also, since the sine function completes one period each stage, it divides the stages equally into alternating values of $\omega$. \\

Now, for the first 0.5 seconds, the angular velocities of the motors are identical allowing the quadcopter to ascend in the first 0.25 seconds and then to decelerate in the next 0.25 seconds until it steadily reaches an approximate of 0.3 m. \\

This is followed by a roll movement, where the values of $\omega_1$ and $\omega_4$ are bigger than the values of $\omega_2$ and $\omega_3$. This rotates the angle $\phi$ to about 12$^{\circ}$ and consequently moves the quadcopter in the y direction to reach approximately -0.1 m in the next 0.5 seconds. The quadcopter will keep moving in the y direction because the angle $\phi$ does not return to zero.\\

Next in the following 0.5 seconds, pitch movement is achieved by increasing the values of $\omega_1$ and $\omega_4$ and decreasing the values of $\omega_2$ and $\omega_3$ which is responsible for rotating the angle $\theta$ of 10$^{\circ}$. Consequently, the quadcopter moves approximately 0.1 m in the x direction. The $\psi$ angles also starts to increase because of the coupling between the quadcopter's states. \\

Finally in the last 0.5 seconds, the yaw movement is achieved by increasing the values of $\omega_1$ and $\omega_3$ and decreasing the values of $\omega_2$ and $\omega_4$ which is responsible for rotating the angle $\psi$ of 10$^{\circ}$. \\

\begin{figure}[h!]
    \centering
    \includegraphics[width=\textwidth]{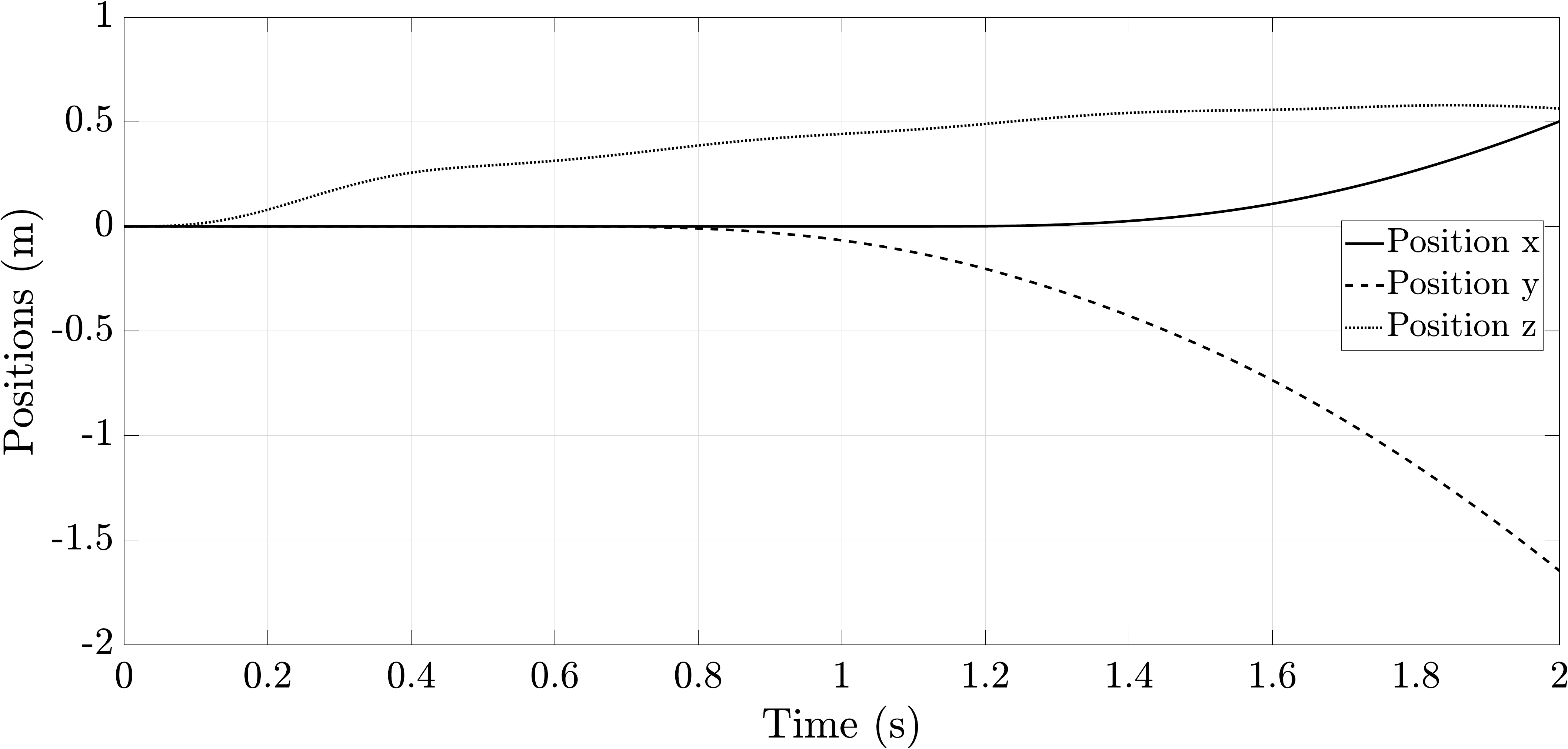}
    \caption{Open-loop positions.}
    \label{fig:openpos}
\end{figure}

\begin{figure}[h!]
    \centering
    \includegraphics[width=\textwidth]{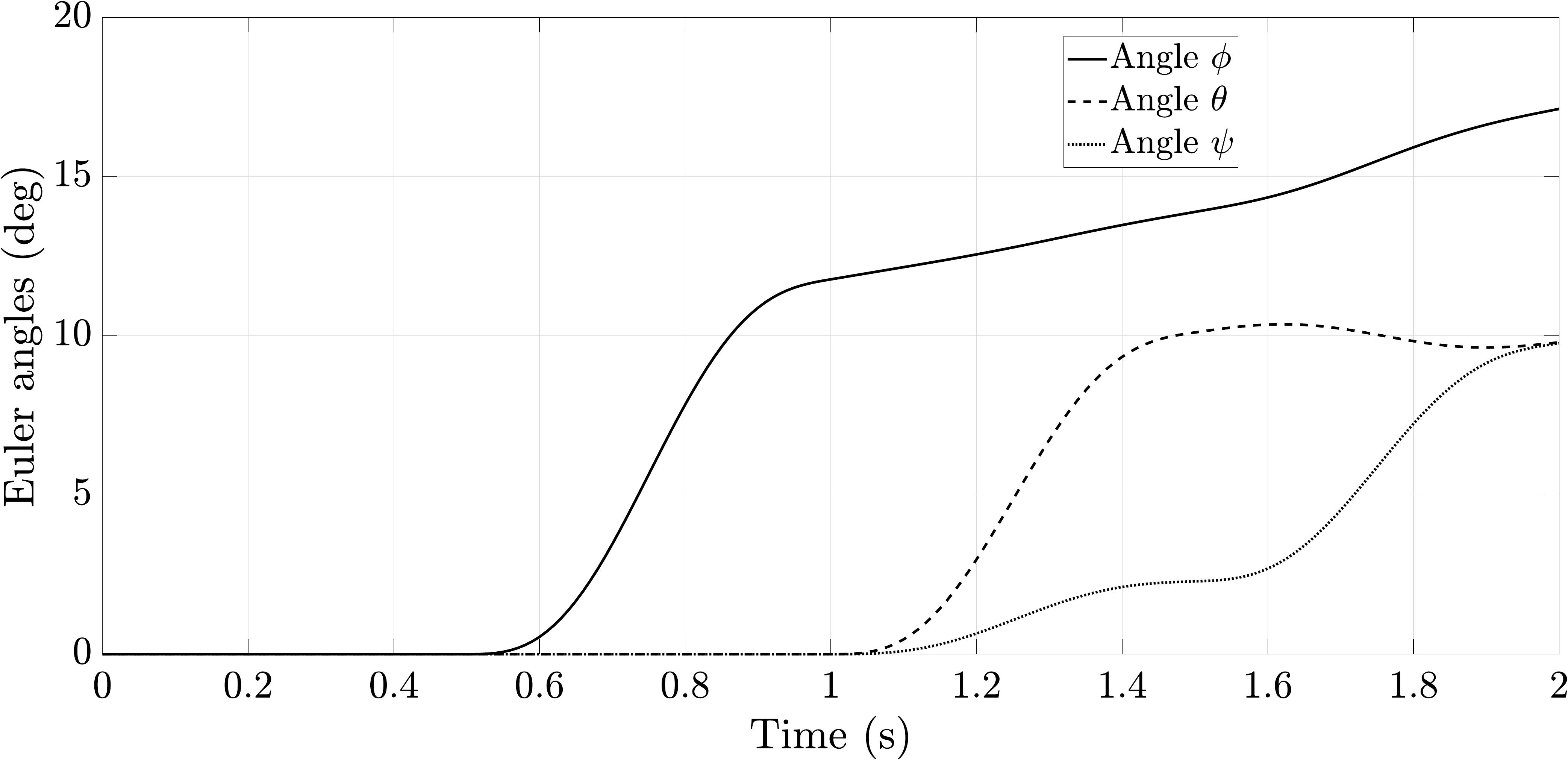}
    \caption{Open-loop Euler angles.}
    \label{fig:openang}
\end{figure}

\begin{figure}[h!]
    \centering
    \includegraphics[width=\textwidth]{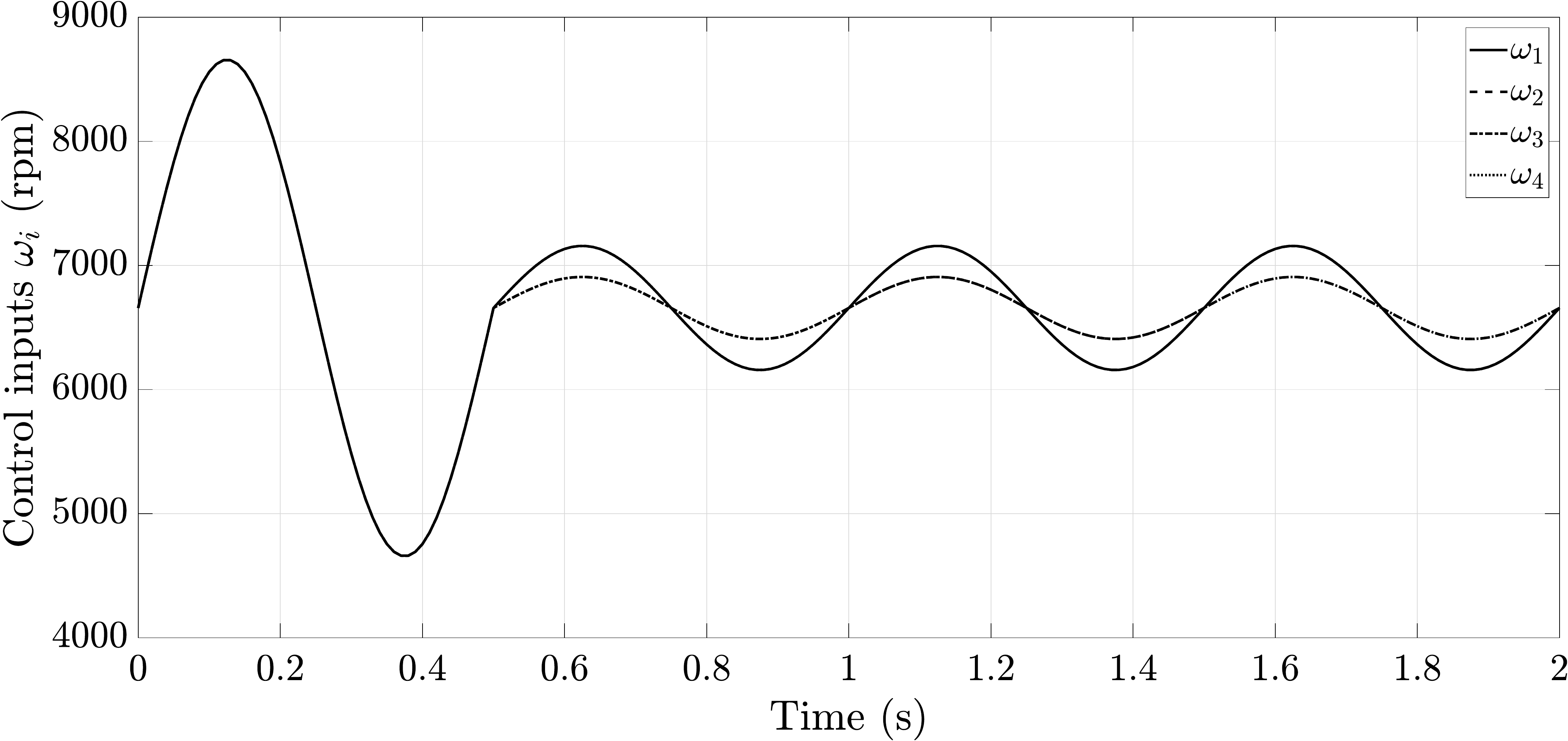}
    \caption{Open-loop control inputs.}
    \label{fig:openinp}
\end{figure}

The inertial positions and angles of the quadcopter during the open-loop simulation are presented in \figurename{ \ref{fig:openpos}} and \figurename{ \ref{fig:openang}} respectively. The control inputs are also illustrated in \figurename{ \ref{fig:openinp}}.
{}

{}

{}



\chapter{Controller design}
\label{ch4}
This chapter presents the controller design approach that is adopted in this work for stabilizing the altitude and attitude of the quadcopter. The controller design is based on a linear approximation of the quadcopter model, which its derivation is explained in details. The controller is then tested on the original system dynamics, the nonlinear system dynamics.\\

The first section of this chapter (\ref{sec:s41} \nameref{sec:s41}) provides a detailed description of the linearization process that is used for control design and also explains the operating point selection criteria. \\

The second section of this chapter (\ref{sec:42} \nameref{sec:42}) provides a detailed description of the various blocks involved in the feedback control loop.\\

The third section of this chapter (\ref{sec:43} \nameref{sec:43}) describes the design process of a PID altitude and attitude controller for the quadcopter. This design process is based on root locus technique by using MATLAB software.\\

The forth section of this chapter (\ref{sec:44} \nameref{sec:44}) provides the response of the nonlinear system dynamics due to the closed-loop system after applying the developed controller.

\section{Linearization}
\label{sec:s41}

Linearization, in short, is a process of approximating the behavior of a nonlinear function by a linear function that is close enough to the original function. It is a technique used in many areas of engineering, notably in control theory, to approximate nonlinearities. The approximation is then used as a basis for the controller design to produce the desired system behaviour.\\

In general, linearization is performed around an operating point that resembles the desired system behaviour. In the case of the quadrotor, the most convenient operating point is the point at which the quadcopter is in hover mode, which means that the quadcopter is at an equilibrium state. This implies the following:
\begin{itemize}
    \item The roll and pitch angles are close to zero, hence Taylor's first order approximation of the trigonometric functions is adopted and thus any higher order terms are omitted.
    \item The angular velocities of the quadcopter are close to zero, hence higher order terms are also omitted.
    \item At an equilibrium state, inertial positions and the yaw movement can be considered constants as they are indifferent in term of the linearization process. 
\end{itemize}{}
At equilibrium state and to make sure the quadcopter stays stationary in the air, the forces generated by the motors should compensate for the weight of the quadcopter, which is created by gravity in the z direction. This means the quadcopter does not accelerate in the z direction, hence $\ddot{z} = 0$. This being said, equation (\ref{eq:zdd}) simplifies to:
\begin{equation}
    F = mg
\end{equation}
where the first order approximation of cos$t$ = 1.\\

Substituting $F$ from equation (\ref{eq:force}):
\begin{equation}
    C_T(\omega_{1e}^2+\omega_{2e}^2+\omega_{3e}^2+\omega_{4e}^2) = mg
\end{equation}{}

Since the quadcopter is assumed to be symmetrical, the angular velocities of the four motors must be equal, this means:
\begin{equation}
    \omega_{1e}^2=\omega_{2e}^2=\omega_{3e}^2=\omega_{4e}^2 = \omega_e^2
\end{equation}{}

This leads to the following equation:
\begin{equation}
\begin{split}
    4C_T \omega_e^2 &= mg, \\[10pt]
    \omega_e &= \sqrt{\frac{mg}{4C_T}}
\end{split}
\end{equation}

Substituting the physical parameters from Table \ref{tab:param}:
\begin{equation}
\label{eq:we}
\begin{split}
    \omega_e &= \sqrt{\frac{0.9 \times 9.81}{4\times 4.980 \times 10^{-8}}},\\[10pt]
    \omega_e &= 6657.5 \text{ [rpm]}
\end{split}
\end{equation}
where $\omega_e$ is the required speed of each motor that would maintain the hover state of the quadcopter. \\

Now, the linear relationship between the angular velocities in the body from and the inertial frame is as follows:
\begin{equation}
    \begin{split}
    \dot{\phi} &= p + \phi \theta q + \theta r\\
    \dot{\theta} &= q- \phi r\\
    \dot{\psi} &= \phi q+r
    \end{split}
\end{equation}
where the first order approximation of sin$t$ and tan$t$ is equal to $t$. \\

By removing all higher order terms, it simplifies to:
\begin{equation}
    \begin{bnmatrix}
    \dot{\phi} \\
    \dot{\theta}\\
    \dot{\psi} 
    \end{bnmatrix}
    =
    \begin{bnmatrix}
    p\\
    q\\
    r
    \end{bnmatrix}
\end{equation}

Finally, the linearized dynamic equations of the quadcopter are:
\begin{equation}
\begin{split}
    \ddot{x} &= \frac{F}{m}(\theta \text{cos}\psi+\phi\text{sin}\psi), \\[10pt]
    \ddot{y}&= \frac{F}{m}(\theta \text{sin}\psi-\phi\text{cos}\psi), \\[10pt]
    \ddot{z} &= \frac{F}{m}-g
\end{split}{}
\quad\quad
\begin{split}
    \ddot{\phi} &= \frac{\tau_\phi}{I_{xx}},\\[10pt]
    \ddot{\theta} &= \frac{\tau_\theta}{I_{xx}},\\[10pt]
    \ddot{\psi} &= \frac{\tau_\psi}{I_{zz}}
\end{split}
\end{equation}

\section{Control system architecture}
\label{sec:42}
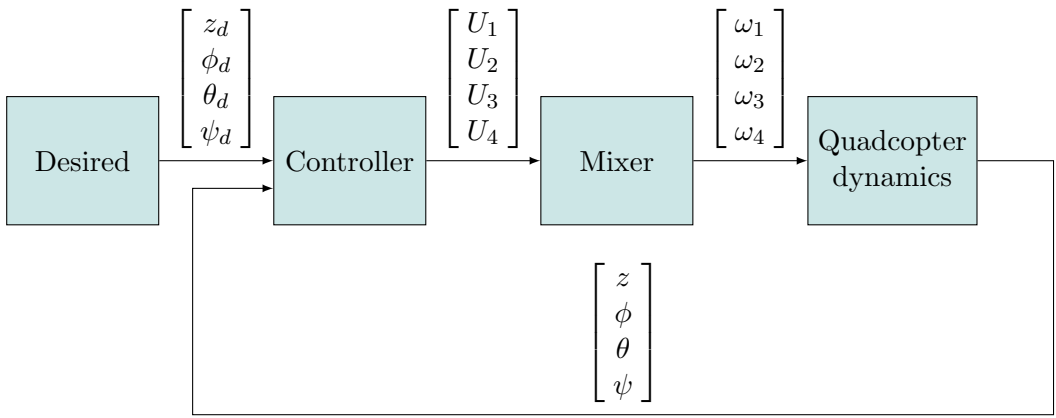
\begin{figure}[H]
    \centering
    \begin{tikzpicture}[auto, node distance=3cm,>=latex,every text node part/.style={align=center}]
    \tikzset{pinstyle/.style={pin edge={to-,thick,black}}}
    \tikzstyle{block} = [draw, fill=teal!20, rectangle, 
    minimum height=1.7cm, minimum width=2cm]
    
    \node [block] (a) {Desired};
    \node [block, right = 1.5 cm of a](b) {Controller};
    \node [block, right = 1.5 cm of b](c) {Mixer};
    \node [block, right = 1.5 cm of c](d) {Quadcopter \\dynamics};
    \node [input, below left= 1.5 cm of b] (feedback)  {};
    \node [input, right= 1 cm of d] (y)  {};

    \draw [->] (a) -- node  {$\begin{bnmatrix}z_d\\\phi_d \\ \theta_d \\ \psi_d \end{bnmatrix}$}(b);
    \draw [->] (b) -- node  {$\begin{bnmatrix}U_1 \\ U_2 \\U_3 \\U_4 \end{bnmatrix}$}(c);
    \draw [->] (c) -- node {$\begin{bnmatrix}\omega_1 \\ \omega_2 \\ \omega_3 \\ \omega_4\end{bnmatrix}$}(d);
    \draw [->] (d) -- (y.south) -| ([yshift=-1.45cm]y |- feedback) -- node[midway,above] {$\begin{bnmatrix}z\\\phi \\ \theta \\ \psi \end{bnmatrix}$} ([yshift=-1.45cm]feedback) |- (b.200);
\end{tikzpicture}
    \caption{Control system architecture.}
    \label{fig:arch_control}
\end{figure}

The architecture of the feedback control system is presented in \figurename{ \ref{fig:arch_control}}.
{}

The system is implemented by four blocks and is designed to allow the quadcopter to keep continuous track of the desired state. The quadcopter dynamics block is explained in details in section (\ref{sec:s31}). The desired block holds the desired target values, which can be either constants or time-varying. \\

The controller block implements four PD controllers for the four desired states. Each controller handles a single desired state. As will be seen in the next section, the controller block output is directly related to the force and torques needed for the quadcopter to track the desired states:
\begin{equation}
    \begin{bnmatrix}
    U_1 \\ U_2 \\ U_3 \\ U_4
    \end{bnmatrix}
    =
    \begin{bnmatrix}
    F \\ \tau_\phi \\ \tau_\theta \\ \tau_\psi
    \end{bnmatrix}
\end{equation}

These control signals need to be converted to he angular velocities of the motors. This is achieved by the mixer block. The mixer block implements the following equations, which are calculated from equation (\ref{eq:force}) and equations (\ref{eq:torques}):
\begin{equation}
    \begin{split}
        \omega_1^2 = \frac{U_1}{4C_T} + \frac{U_2}{4dC_T} + \frac{U_3}{4dC_T} + \frac{U_4}{4C_D}\\[10pt]
        \omega_2^2 = \frac{U_1}{4C_T} - \frac{U_2}{4dC_T} + \frac{U_3}{4dC_T} - \frac{U_4}{4C_D}\\[10pt]
        \omega_3^2 = \frac{U_1}{4C_T} - \frac{U_2}{4dC_T} - \frac{U_3}{4dC_T} + \frac{U_4}{4C_D}\\[10pt]
        \omega_4^2 = \frac{U_1}{4C_T} + \frac{U_2}{4dC_T} - \frac{U_3}{4dC_T} - \frac{U_4}{4C_D}
    \end{split}
\end{equation}

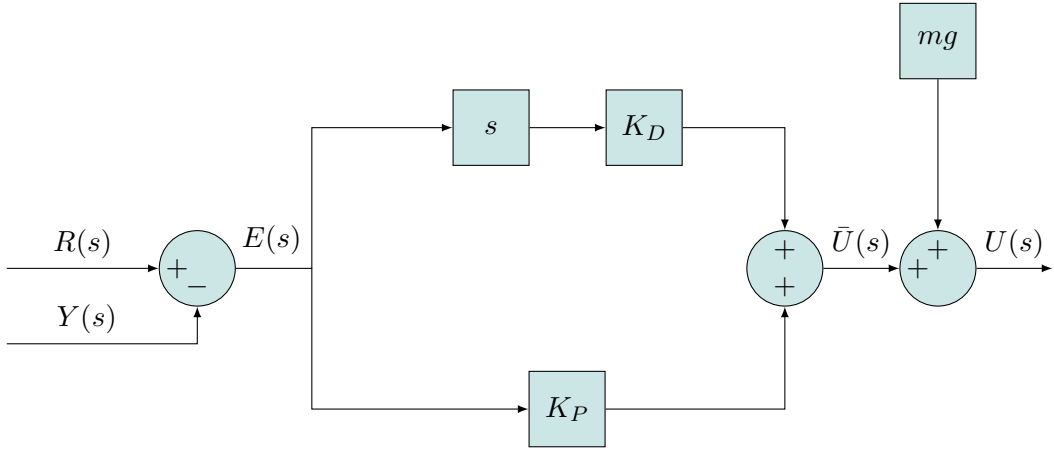
\begin{figure}[h!]
    \centering
    \begin{tikzpicture}[auto, node distance=3cm,>=latex,every text node part/.style={align=center}]
    \tikzstyle{block} = [draw, fill=teal!20, rectangle, 
    minimum height=1cm, minimum width=1cm]
    \tikzstyle{sum} = [draw, fill=teal!20, circle,minimum size=1cm, node distance=1cm]
    \node [input, name=input] {};
    \node[input, below = 1 cm of input] (input1) {};
    \node [sum, right = 2 cm of input] (sum1) {};
    \node [block, above right =  1cm and 3 cm of sum1] (deriv) {$s$};
    \node [block, right = 1 cm of deriv] (kp) {$K_D$};
    \node [block, below right =  1cm and 4 cm of sum1] (ki) {$K_P$};
    \node [sum,  above right  = 1 cm and 2 cm of ki] (sum2) {};
    \node [sum,  right  = 1 cm of sum2] (sum3) {};
    \node[block, above = 2 cm of sum3] (input2) {$mg$};
    \node [output, right=1 cm of sum3] (output) {};
    
    \node[left=-15pt of sum1] {$+$};
    \node[below=-15pt of sum1] {$-$};
    \node[above=-15pt of sum2] {$+$};
    \node[below=-15pt of sum2]  {$+$};
    \node[above=-15pt of sum3] {$+$};
    \node[left=-15pt of sum3]  {$+$};
    
    \draw [->] (input) -- node{$R(s)$} (sum1);
    \draw [->] (input1) -| node[pos=0.21,above]{$Y(s)$} (sum1);
    \draw [->] (sum1) -- ([xshift=1 cm]sum1.east) |- node[pos=0.1]{$E(s)$} (deriv.west);
    \draw [->] ([xshift=1 cm]sum1.east) |- (ki.west);
    \draw [->] (deriv) -- (kp);
    \draw [->] (ki) -| (sum2);
    \draw [->] (kp) -| (sum2);
    \draw [->] (input2) -- (sum3);
    \draw [->] (sum2) -- node {$\bar{U} (s)$}(sum3);
    \draw [->] (sum3) -- node {$U (s)$}(output);
    
\end{tikzpicture}
    \caption{Altitude controller internal architecture.}
    \label{fig:arch_internalaltcontrol}
\end{figure}

Back to the controller block, \figurename{ \ref{fig:arch_internalcontrol}} shows the control loop architecture for the attitude controller while \figurename{ \ref{fig:arch_internalaltcontrol}} shows the control loop architecture for the altitude controller. 
\begin{figure}[h!]
    \centering
    \begin{tikzpicture}[auto, node distance=3cm,>=latex,every text node part/.style={align=center}]
    \tikzstyle{block} = [draw, fill=teal!20, rectangle, 
    minimum height=1cm, minimum width=1cm]
    \tikzstyle{sum} = [draw, fill=teal!20, circle,minimum size=1cm, node distance=1cm]
    \node [input, name=input] {};
    \node[input, below = 1 cm of input] (input1) {};
    \node [sum, right = 2 cm of input] (sum1) {};
    \node [block, above right =  1cm and 3 cm of sum1] (deriv) {$s$};
    \node [block, right = 1 cm of deriv] (kp) {$K_D$};
    \node [block, below right =  1cm and 4 cm of sum1] (ki) {$K_P$};
    \node [sum,  above right  = 1 cm and 2 cm of ki] (sum2) {};
    \node [output, right=1 cm of sum2] (output) {};
    
    \node[left=-15pt of sum1] (aaa) {$+$};
    \node[below=-15pt of sum1] (aaaa) {$-$};
    \node[above=-15pt of sum2] (a1aa) {$+$};
    \node[below=-15pt of sum2] (a1aaa) {$+$};
    
    \draw [->] (input) -- node{$R(s)$} (sum1);
    \draw [->] (input1) -| node[pos=0.21,above]{$Y(s)$} (sum1);
    \draw [->] (sum1) -- ([xshift=1 cm]sum1.east) |- node[pos=0.1]{$E(s)$} (deriv.west);
    \draw [->] ([xshift=1 cm]sum1.east) |- (ki.west);
    \draw [->] (deriv) -- (kp);
    \draw [->] (ki) -| (sum2);
    \draw [->] (kp) -| (sum2);
    \draw [->] (sum2) -- node {$U(s)$}(output);
    
\end{tikzpicture}
    \caption{Attitude controller internal architecture.}
    \label{fig:arch_internalcontrol}
\end{figure}{}

{}
where $R(s)$ is the desired input, $Y(s)$ is the current output, $E(s)$ is the error signal, $U(s)$ is the control signal, block "$s$" represents the derivation operation, $K_P$ is the proportional coefficient and $K_D$ is the derivative coefficient. \\

The reason behind adding a forward term, $mg$, in the altitude control loop and discarding the integral term in both loops is discussed in the next section.

\section{Control system design}
\label{sec:43}
After the mathematical model of the quadcopter was linearized, it is time to transform the equations, which need to be controlled, to the $s$ domain. For the linearized Euler equations, this transformation is straightforward and is given as follows:

\begin{equation}
    \begin{split}
        \frac{\phi(s)}{\tau_\phi} =\frac{\phi(s)}{U_2(s)} &= \frac{1}{I_{xx}s^2}\\[10pt]
        \frac{\theta(s)}{\tau_\theta} =\frac{\theta(s)}{U_3(s)} &= \frac{1}{I_{yy}s^2}\\[10pt]
        \frac{\psi(s)}{\tau_\psi} =\frac{\psi(s)}{U_4(s)} &= \frac{1}{I_{zz}s^2}
    \end{split}
\end{equation}

The altitude equation does not satisfy the linear mapping conditions, which is discussed in Appendix (\ref{app2}). Therefore, it cannot be directly transformed to a transfer function. A workaround to this problem is as follows:
\begin{enumerate}
    \item Rearrange the equation:
    \begin{equation}
        m\ddot{z} = F - mg
    \end{equation}
    \item Make the following substitution:
    \begin{equation}
        \bar{U}_1 = U_1 - mg, \quad U_1=F
    \end{equation}
    \item The equation becomes:
    \begin{equation}
        m\ddot{z} = \bar{U}_1
    \end{equation}
    \item Transform the equation to the $s$ domain:
    \begin{equation}
        \frac{z(s)}{\bar{U}_1} = \frac{1}{ms^2}
    \end{equation}
    \item Make sure to substitute $U_1$ back to the equation after the controller is designed as shown in \figurename{ \ref{fig:arch_internalaltcontrol}}.
\end{enumerate}

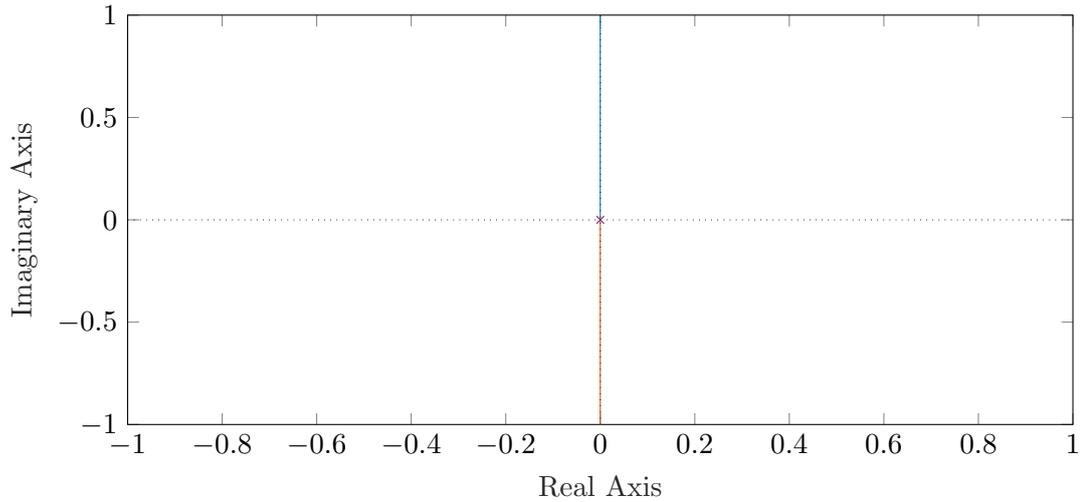
\begin{figure}[h!]
    \centering
%
%
\definecolor{mycolor1}{rgb}{0.00000,0.44700,0.74100}%
\definecolor{mycolor2}{rgb}{0.85000,0.32500,0.09800}%
\definecolor{mycolor3}{rgb}{0.92900,0.69400,0.12500}%
\definecolor{mycolor4}{rgb}{0.49400,0.18400,0.55600}%
\begin{tikzpicture}

\begin{axis}[%
width=0.983\fwidth,
height=0.429\fwidth,
at={(0\fwidth,0\fwidth)},
scale only axis,
unbounded coords=jump,
xmin=-1,
xmax=1,
xlabel style={font=\color{white!15!black}},
xlabel={Real Axis},
ymin=-1,
ymax=1,
ylabel style={font=\color{white!15!black}},
ylabel={Imaginary Axis},
axis background/.style={fill=white},
legend style={legend cell align=left, align=left, draw=white!15!black}
]
\addplot [color=mycolor1]
  table[row sep=crcr]{%
0	0\\
-0	0.01\\
-0	0.0112344934402052\\
-0	0.0126213842858013\\
-0	0.0141794858965143\\
-0	0.0159299341289871\\
-0	0.0178964740475005\\
-0	0.0201057820289447\\
-0	0.0225878276314373\\
-0	0.0253762801353867\\
-0	0.028508965271781\\
-0	0.032028378333286\\
-0	0.0359822606285711\\
-0	0.0404242470995434\\
-0	0.0454145938865052\\
-0	0.0510209957107523\\
-0	0.0573195041625182\\
-0	0.0643955593509622\\
-0	0.0723451489106727\\
-0	0.0812761100867617\\
-0	0.0913095925615115\\
-0	0.102581701866011\\
-0	0.115245345669878\\
-0	0.129472307994241\\
-0	0.145455579484953\\
-0	0.163411975356494\\
-0	0.18358507651935\\
-0	0.206248533787619\\
-0	0.231709779988894\\
-0	0.260314200331661\\
-0	0.292449817601829\\
-0	0.328552555743694\\
-0	0.369112153226517\\
-0	0.41467880641233\\
-0	0.465870633043142\\
-0	0.52338205709074\\
-0	0.587993228710699\\
-0	0.66058060708354\\
-0	0.742128849700676\\
-0	0.833744169374924\\
-0	0.936669340165187\\
-0	1.05230055577271\\
-0	1.18220636909527\\
-0	1.32814896985696\\
-0	1.49210808894732\\
-0	1.67630785373557\\
-0	1.88324695865567\\
-0	2.11573256033034\\
-0	2.37691835702596\\
-0	2.67034736899114\\
-0	3\\
-0	60\\
};

\addplot [color=mycolor2]
  table[row sep=crcr]{%
-0	0\\
0	-0.01\\
0	-0.0112344934402052\\
0	-0.0126213842858013\\
0	-0.0141794858965143\\
0	-0.0159299341289871\\
0	-0.0178964740475005\\
0	-0.0201057820289447\\
0	-0.0225878276314373\\
0	-0.0253762801353867\\
0	-0.028508965271781\\
0	-0.032028378333286\\
0	-0.0359822606285711\\
0	-0.0404242470995434\\
0	-0.0454145938865052\\
0	-0.0510209957107523\\
0	-0.0573195041625182\\
0	-0.0643955593509622\\
0	-0.0723451489106727\\
0	-0.0812761100867617\\
0	-0.0913095925615115\\
0	-0.102581701866011\\
0	-0.115245345669878\\
0	-0.129472307994241\\
0	-0.145455579484953\\
0	-0.163411975356494\\
0	-0.18358507651935\\
0	-0.206248533787619\\
0	-0.231709779988894\\
0	-0.260314200331661\\
0	-0.292449817601829\\
0	-0.328552555743694\\
0	-0.369112153226517\\
0	-0.41467880641233\\
0	-0.465870633043142\\
0	-0.52338205709074\\
0	-0.587993228710699\\
0	-0.66058060708354\\
0	-0.742128849700676\\
0	-0.833744169374924\\
0	-0.936669340165187\\
0	-1.05230055577271\\
0	-1.18220636909527\\
0	-1.32814896985696\\
0	-1.49210808894732\\
0	-1.67630785373557\\
0	-1.88324695865567\\
0	-2.11573256033034\\
0	-2.37691835702596\\
0	-2.67034736899114\\
0	-3\\
0	-60\\
};

\addplot [color=mycolor3, draw=none, mark=x, mark options={solid, mycolor3}]
  table[row sep=crcr]{%
0	0\\
};

\addplot [color=mycolor4, draw=none, mark=x, mark options={solid, mycolor4}]
  table[row sep=crcr]{%
0	0\\
};

\addplot [color=black, dotted, line width=0.3pt]
  table[row sep=crcr]{%
0	-1\\
0	1\\
};

\addplot [color=black, dotted, line width=0.3pt]
  table[row sep=crcr]{%
-1	0\\
1	0\\
};

\end{axis}
\end{tikzpicture}%
    \caption{Root locus of the open-loop transfer functions.}
    \label{fig:nocontrollerlocus}
\end{figure}

It is obvious that the transfer functions exhibit double integrator behaviour which means that the steady state error is zero, as explained in Appendix \ref{app3}, thus the integral term in the controller is not required. This also means the system is unstable due to the fact that two poles are on the jw-axis, as shown in the root locus plot in \figurename{ \ref{fig:nocontrollerlocus}}. \\

\begin{figure}[h!]
    \centering
    \input{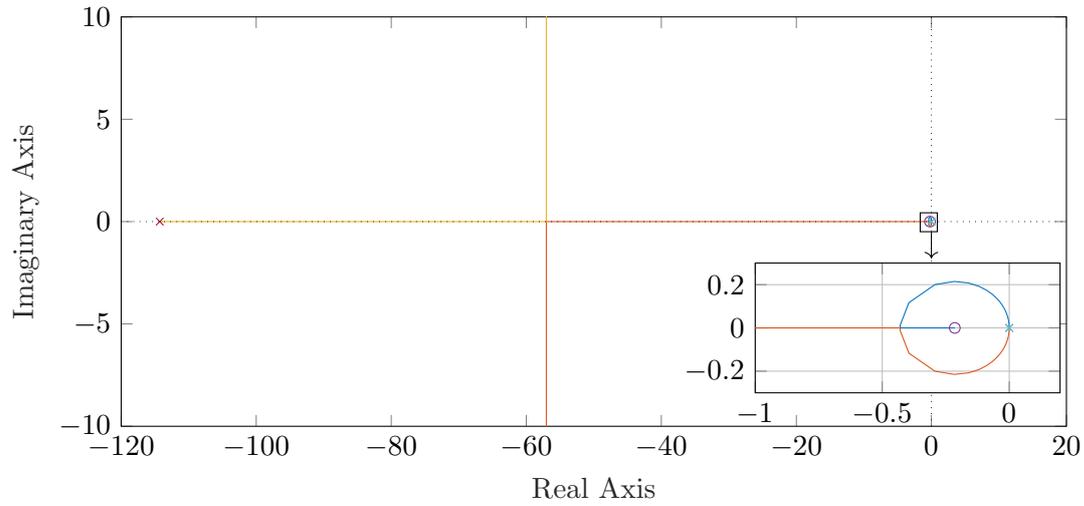}
    \caption{Root locus of the controlled altitude and attitude loop.}
    \label{fig:zlocus}
\end{figure}

In this work, root locus design technique is adopted to design the PD controller with the help of the Root Locus Designer that is provided by the MATLAB software. \figurename{ \ref{fig:zlocus}} shows the reshaped root locus for the controlled altitude and attitude loop with the following PD controller parallel form:

\begin{equation}
u(s) = K_p+K_d\frac{N}{1+N\frac{1}{s}}
\end{equation}
where $N$ is the filter coefficient.\\

The coefficient values of the four PD controller are presented in Table \ref{tab:paramPD}. The step response characteristics of these values is described in Table \ref{tab:stepres}.

\begin{table}[h]
    \centering
    \caption{Coefficient values for the PD controllers.}
    \label{tab:paramPD}
    \setlength{\extrarowheight}{.95ex}
    \begin{tabular}{|>{\columncolor{Gray}}c|c|c|c|}
        \toprule
        \rowcolor{Gray}
         \textbf{Parameter} & \textbf{Proportional gain $K_p$} & \textbf{Derivative gain $K_d$} & \textbf{Filter value $N$}  \\ 
         \midrule
         $z$ & $1.89\times 10^{-1}$ & $8.78\times 10^{-1}$ & 114.286 \\
         $\phi$ & $3.08\times10^{-3}$ & $1.43 \times 10^{-2}$ & 114.286 \\
         $\theta$ & $3.08\times10^{-3}$ & $1.43 \times 10^{-2}$ & 114.286 \\
         $\psi$ & $3.08\times10^{-3}$ & $1.43 \times 10^{-2}$ & 114.286 \\[5pt]
         \bottomrule
    \end{tabular}
\end{table}{}

\begin{table}[h!]
    \centering
    \caption{Step response characteristics of the PD controllers.}
    \label{tab:stepres}
    \setlength{\extrarowheight}{.95ex}
    \begin{tabular}{|>{\columncolor{Gray}}l|cc|>{\columncolor{Gray}}l|c|}
        \toprule
        \rowcolor{Gray}
         \multicolumn{1}{|l|}{\textbf{Performance}} & \textbf{Value} && \multicolumn{1}{l|}{\textbf{Performance}} & \textbf{Value}\\ 
         \midrule
         Rise time & 1.53 seconds && Settling time & 12 seconds  \\
         Overshoot & 12.5\% && Peak & 1.12\\
         Closed-loop stability & Stable &&&\\[5pt]
         \bottomrule
    \end{tabular}
\end{table}{}

\section{Closed-loop behaviour}
\label{sec:44}
In this section, the performance of the developed PD controllers is tested by a case study. The case study is intended to demonstrate the ability of the developed controllers to stabilize the quadcopter from an initial conditions. \\

The developed feedback loop, presented in \figurename{ \ref{fig:arch_control}}, is implemented in MATLAB Simulink 2019a using the PD values obtained in Table \ref{tab:paramPD}. The initial conditions given to the quadcopter are:
\begin{equation}
    \begin{bnmatrix}
    z_i \\ \phi_i \\ \theta_i \\ \psi_i
    \end{bnmatrix}
    =
    \begin{bnmatrix}
    2 \text{ [m]} \\ 15^\circ\\ 15^\circ\\ 15^\circ
    \end{bnmatrix}
\end{equation}

\begin{figure}[h!]
    \centering
\includegraphics[width=\textwidth]{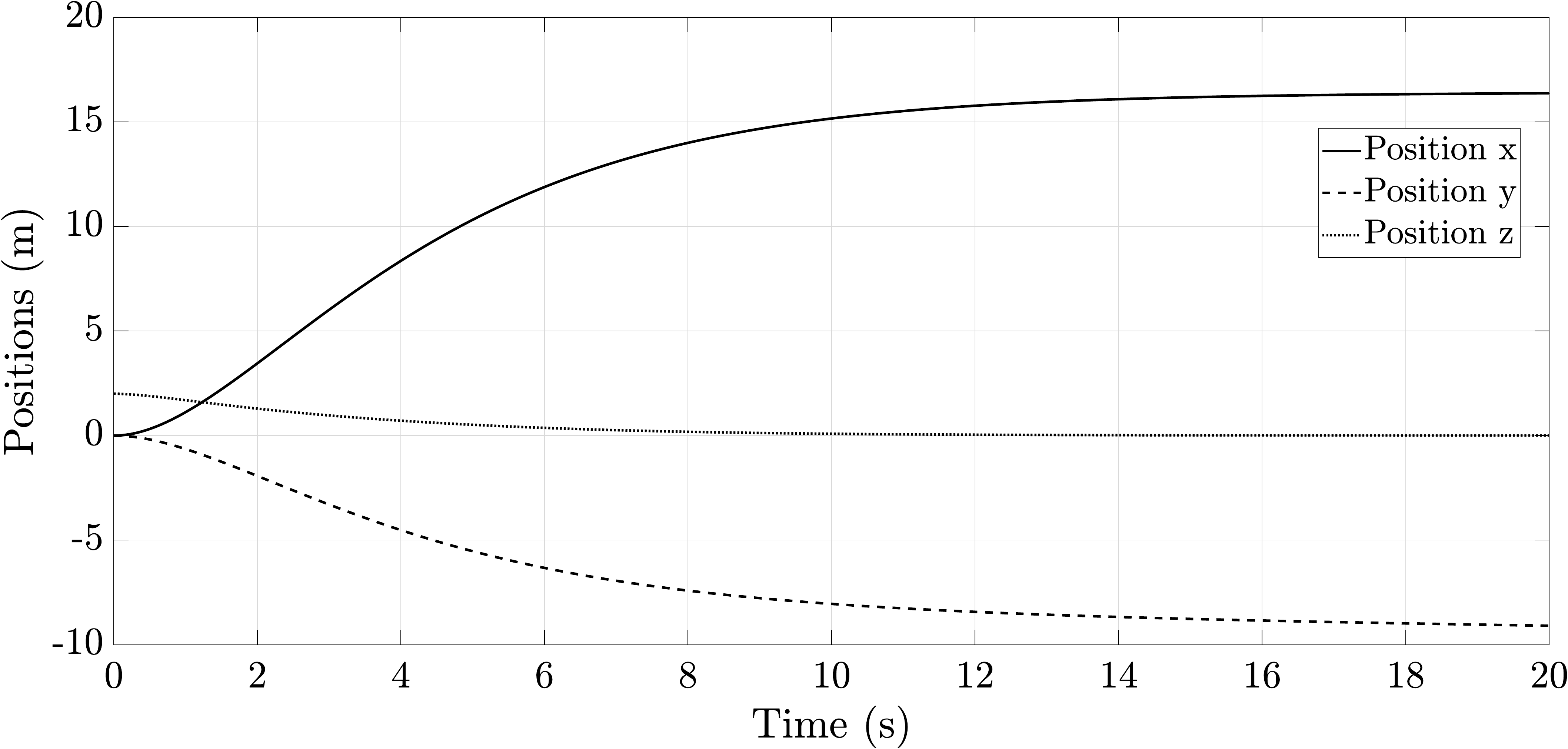}
    \caption{Closed-loop positions.}
    \label{fig:closepos}
\end{figure}

\begin{figure}[h!]
    \centering
    \includegraphics[width=\textwidth]{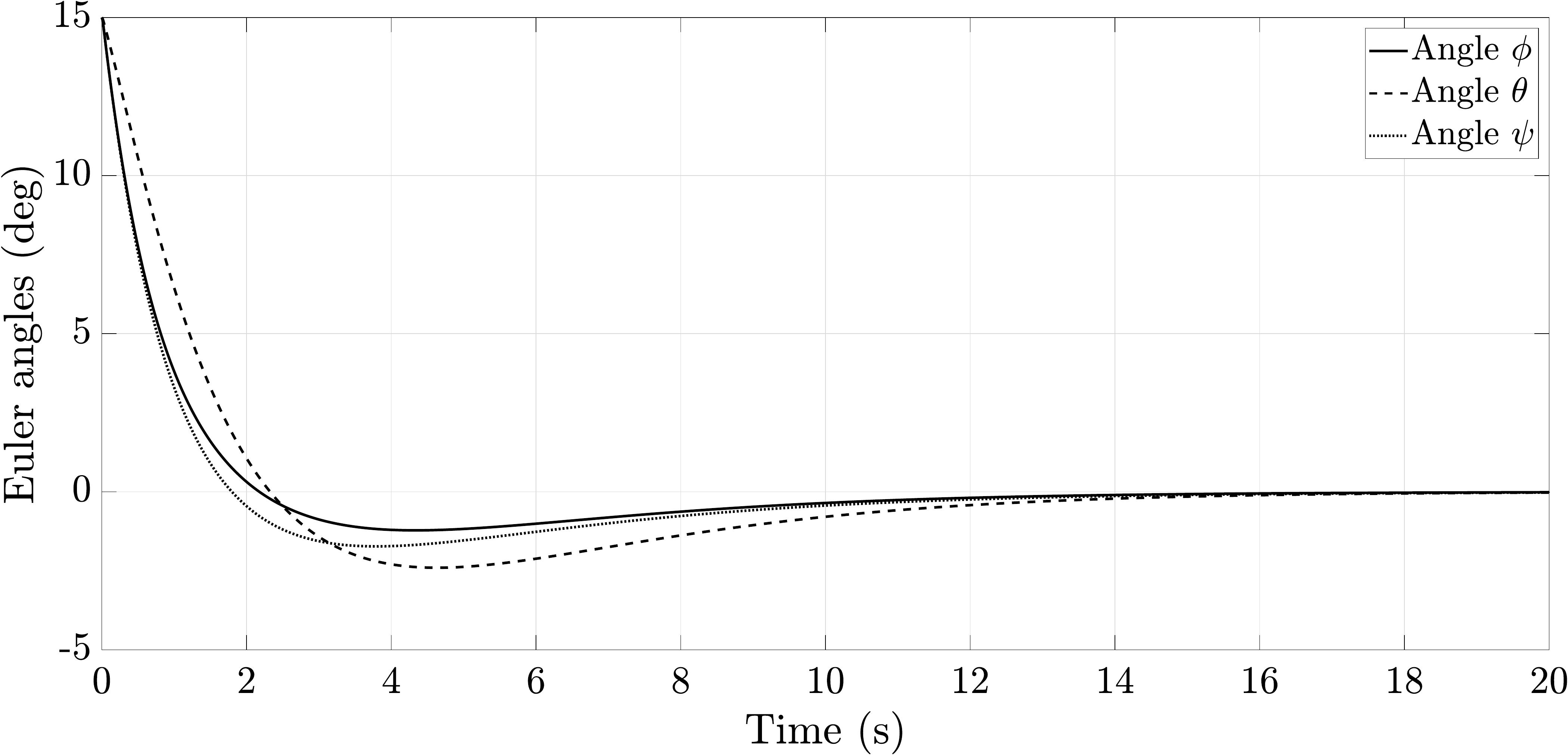}
    \caption{Closed-loop Euler angles.}
    \label{fig:closeang}
\end{figure}

\begin{figure}[h!]
    \centering
    \includegraphics[width=\textwidth]{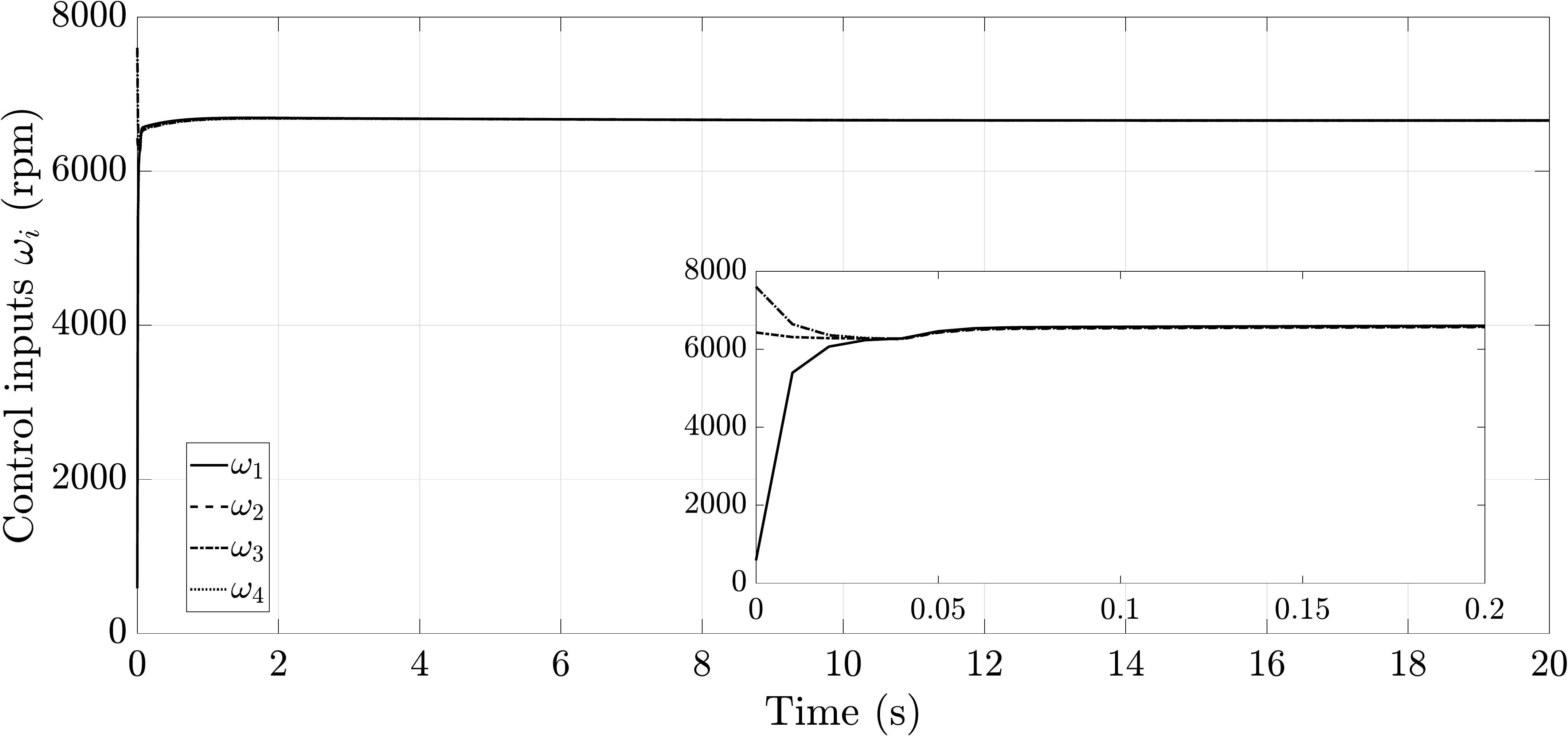}
    \caption{Closed-loop control signals.}
    \label{fig:closeinp}
\end{figure}

The desired state for the altitude and the attitude is zero, since the quadcopter is required to maintain the hover state. As before, the simulation progress at a sample time of 0.01 seconds to a total of 20 seconds. The position, attitude and control signals during the simulation are illustrated in \figurename{ \ref{fig:closepos}}, \figurename{ \ref{fig:closeang}} and \figurename{ \ref{fig:closeinp}} respectively. \\

{}

{}

{}

As expected, the attitude reaches the desired value in about 12 seconds while the altitude takes less time of about 6 seconds to reach zero. The attitude initial values causes the x position to settle at round 16 m and the y position at around -10 m. \\

The control signals shows a smooth response with maximum value of about 8000 rpm and a steady state value of about 6600 rpm. This makes sure that the response is feasible and can be implemented in a hardware with fine tuning. 

\chapter{Advanced Stabilization Controller}
\label{ch6}

\section{Fuzzy inference system}
Fuzzy logic reflects how humans reason about changes, as it represents a continues small changes to an output based on a given input. This is a natural extension of the simple logic (e.g. finite state machines, Boolean) to include more powerful operations such as the ability to represent a continues set of states and outputs. \\

This means that for example representing a change in temperature, $f(T)$, using Boolean logic would be something like this:
\begin{equation}
    f(T) = 
\begin{cases}
0, \quad 0 \leq T \leq 10 \\
1, \quad 10 < T \leq 20
\end{cases}{}
\end{equation}

The conventional logic uses sharp distinctions between cases of a class. For some applications, this is considered a problem because the output sharply changes even if the input changes in a small amount (e.g. from 10 to 10.1). Fuzzy logic solves exactly this problem. \\

The working of fuzzy inference system mainly consists of three steps, which are briefly as follows:
\begin{enumerate}
    \item Fuzzification that is used to map crisp inputs into corresponding fuzzy values based on predefined functions, called membership functions.
    \item Rule evaluation that is used to evaluate predefined rules, which are IF-THEN statements, based on given fuzzy values.
    \item Defuzzification which is used to convert a fuzzy output to a crisp one.
\end{enumerate}

In this work, these steps are achieved with the help of Fuzzy Toolbox of MATLAB software. Also, the membership functions and the rules are obtained automatically as will be discussed next. For more details of the fuzzy inference system, check the book by \citet{passino1998fuzzy}.

\section{ANFIS controller}
Designing a fuzzy controller can be rather a hard task and for most of the times, an expert is required to define precise rules and shape functional membership functions. Adaptive Neuro-Fuzzy Inference System (ANFIS) is considered a more efficient and optimal way to construct such controller. ANFIS uses the power of neural network to construct an optimized fuzzy inference system controller.\\

Training data that consists of the error and its rate of change are collected and fed to the neural network along with the controller output to train and construct reliable fuzzy controller. Fuzzy Toolbox of MATLAB is used to develop the controller due to its ease of use. The inputs and outputs for the four controllers are presented in Table \ref{tab:fuzzyinput}. It is to be noted that roll, pitch and yaw movements share the same fuzzy controller .

\begin{table}[h!]
    \centering
    \caption{The inputs and output of the fuzzy controller.}
    \label{tab:fuzzyinput}
    \setlength{\extrarowheight}{.95ex}
    \begin{tabular}{|>{\columncolor{Gray}}l|c|c|c|}
        \toprule
        \rowcolor{Gray}
         \textbf{Controller} & \textbf{Input 1} & \textbf{Input 2} & \textbf{Output}\\ 
         \midrule
         Altitude & $e_z$ & $\dot{e}_z$ & $U_1$ \\
         Attitude & $e_{\phi,\theta,\psi}$ &$\dot{e}_{\phi,\theta,\psi}$ & $U_{2,3,4}$ \\
         \bottomrule
    \end{tabular}
\end{table}{}
where $e_t$ is the error between the desired state, $t_{d}$, and the actual state, $t_{a}$. The error is calculated as follows:
\begin{equation}
    e_t = t_d - t_a
\end{equation}

\begin{figure}[h]
    \centering
     \begin{subfigure}[b]{0.49\textwidth}
         \centering
         \includegraphics[width=\textwidth]{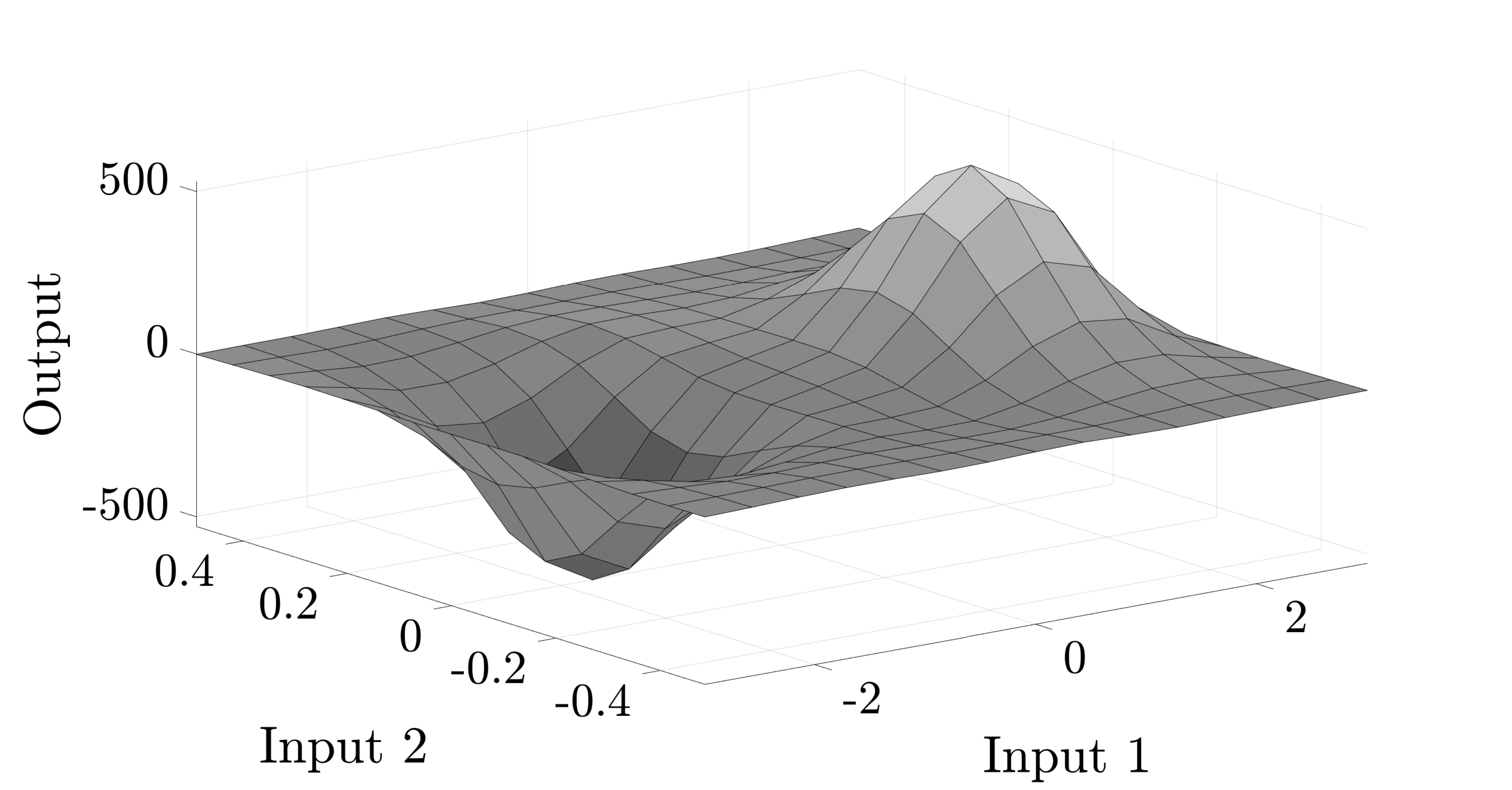}
         \caption{Altitude}
     \end{subfigure}
     \begin{subfigure}[b]{0.49\textwidth}
         \centering
         \includegraphics[width=\textwidth]{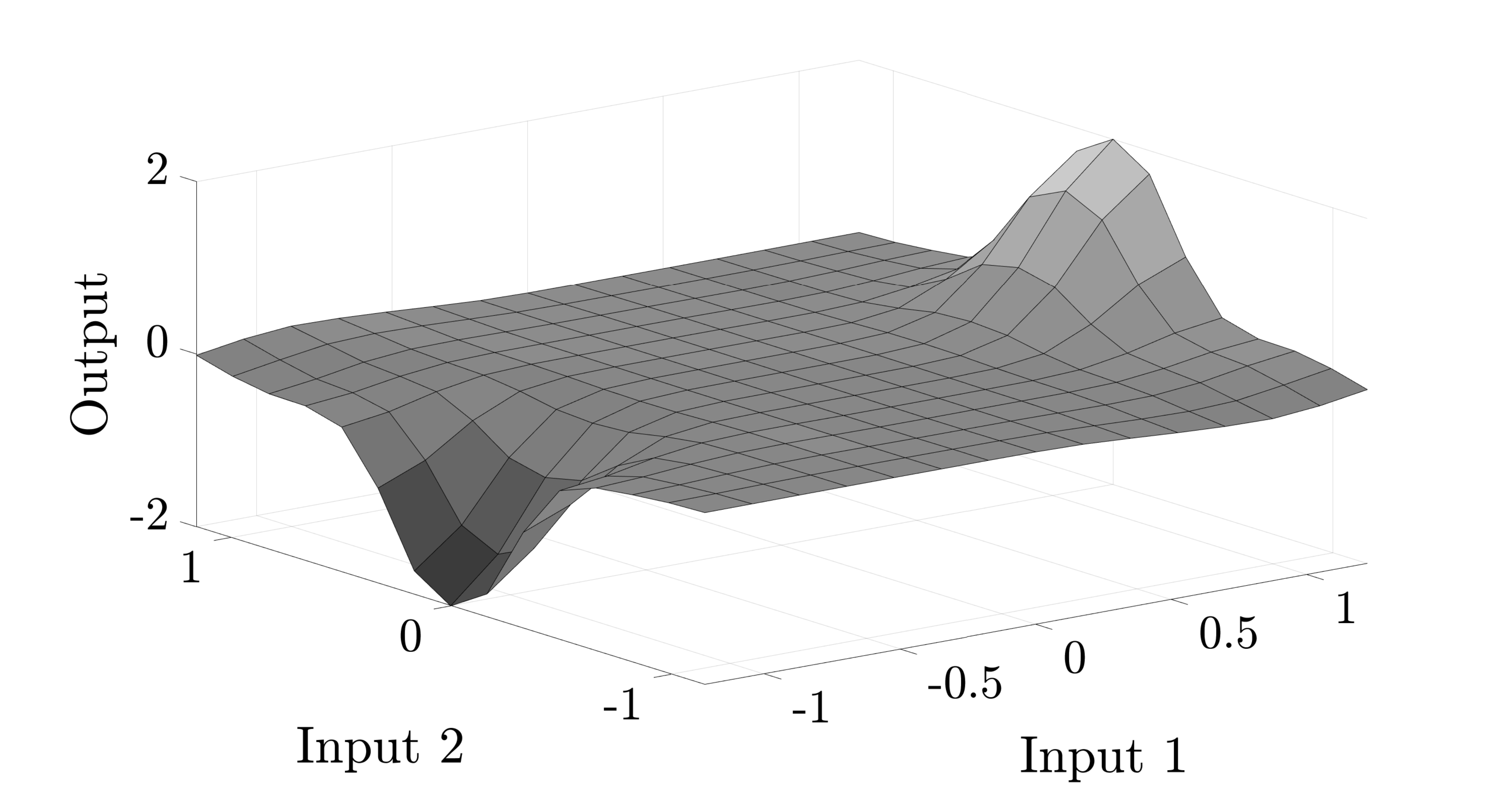}
         \caption{Attitude}
     \end{subfigure}
    \caption{Surface rules of the fuzzy controller.}
    \label{fig:surf_fuzzy}
\end{figure}

\begin{figure}[H]
    \centering
     \begin{subfigure}[b]{0.49\textwidth}
         \centering
         \includegraphics[width=\textwidth]{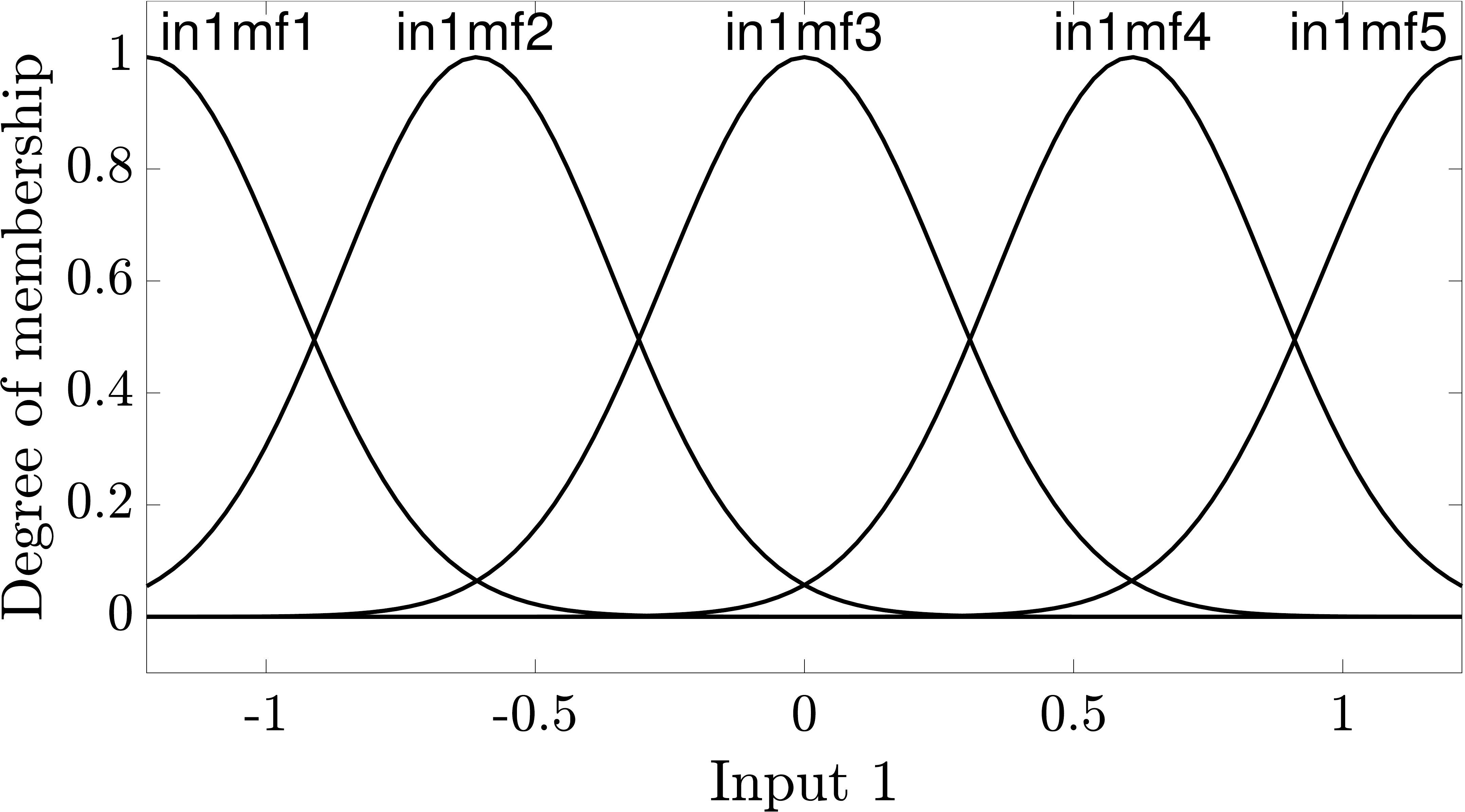}
         \caption{Attitude: Input 1}
     \end{subfigure}
     \begin{subfigure}[b]{0.49\textwidth}
         \centering
         \includegraphics[width=\textwidth]{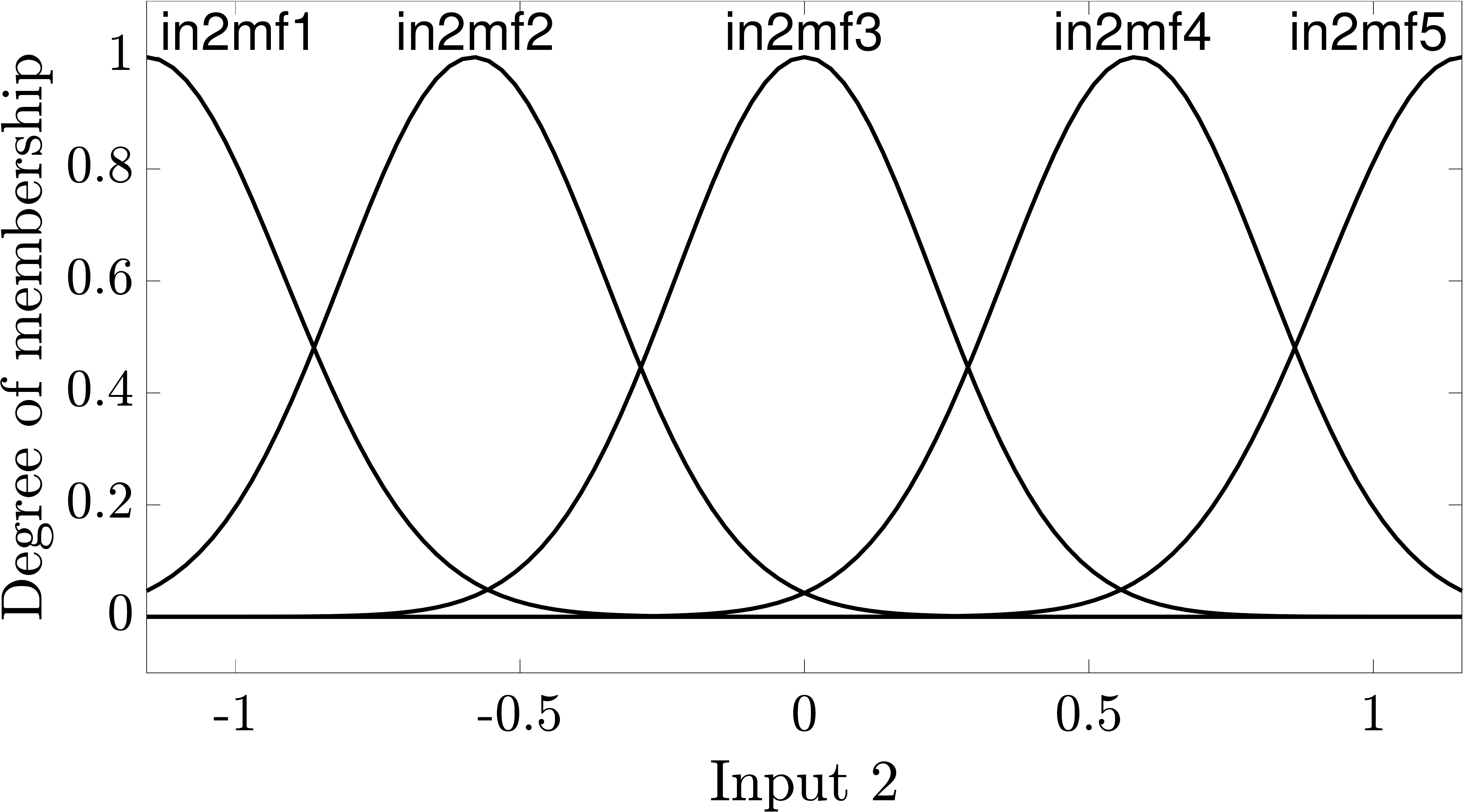}
         \caption{Attitude: Input 2}
     \end{subfigure}
     \begin{subfigure}[b]{0.49\textwidth}
         \centering
         \includegraphics[width=\textwidth]{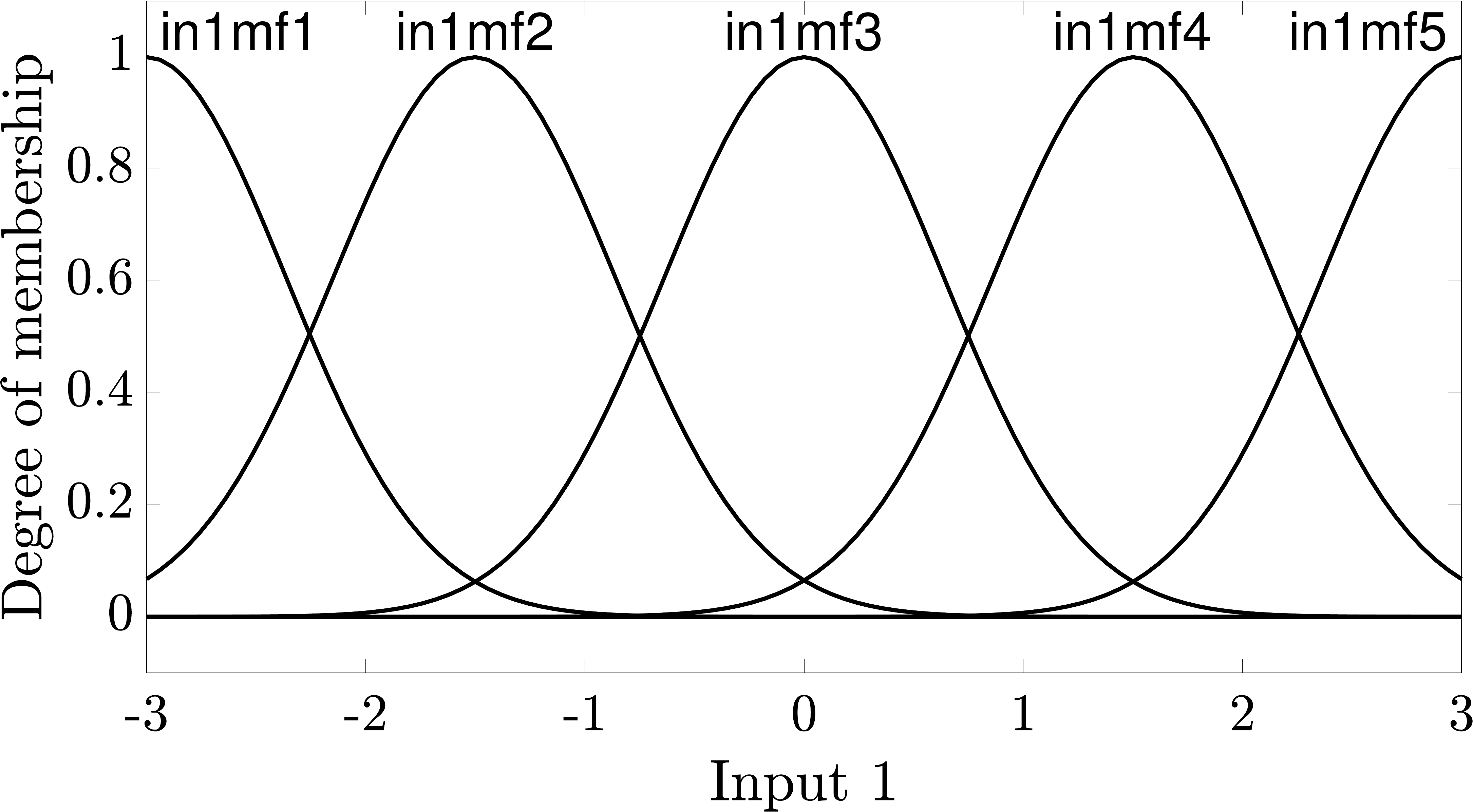}
         \caption{Altitude: Input 1}
     \end{subfigure}
     \begin{subfigure}[b]{0.49\textwidth}
         \centering
         \includegraphics[width=\textwidth]{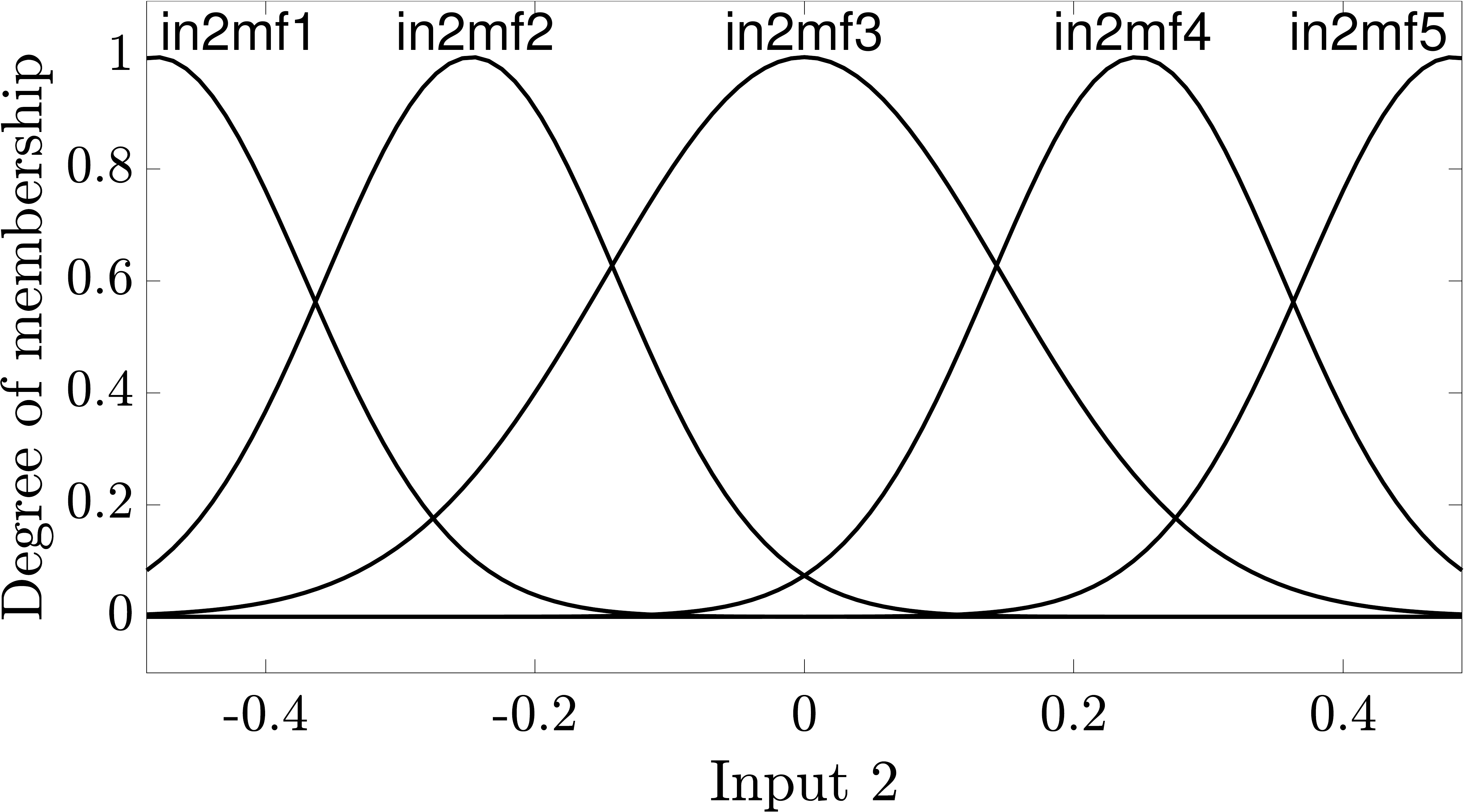}
         \caption{Altitude: Input 2}
     \end{subfigure}
    \caption{Membership functions of the fuzzy controller.}
    \label{fig:mb_fuzzy}
\end{figure}

After the training is complete, five membership functions and a set of rules are constructed. \figurename{ \ref{fig:surf_fuzzy}} shows the rules expressed in a surface form for both attitude and altitude while \figurename{ \ref{fig:mb_fuzzy}} shows the generated membership functions.

{}

\section{Closed-loop response}
In this section, the performance of the developed fuzzy controller is compared to the PD controller via two cases. The cases are intended to demonstrate the ability of the developed fuzzy controllers to stabilize the quadcopter from an initial conditions. \\

The developed feedback loop, presented in \figurename{ \ref{fig:arch_control}}, is implemented in MATLAB Simulink 2019a using the fuzzy controller membership functions presented in \figurename{ \ref{fig:mb_fuzzy}}. The desired state for the altitude and the attitude is zero for both cases, since the quadcopter is required to maintain the hover state. As before, the simulation progress at a sample time of 0.01 seconds to a total of 20 seconds.

\subsection{Case I}
For the second case, the initial conditions given to the quadcopter are:
\begin{equation}
    \begin{bnmatrix}
    z_i \\ \phi_i \\ \theta_i \\ \psi_i
    \end{bnmatrix}
    =
    \begin{bnmatrix}
    2 \text{ [m]} \\ 30^\circ\\ -30^\circ\\ 0^\circ
    \end{bnmatrix}
\end{equation}

\begin{figure}[h!]
    \centering
     \begin{subfigure}[b]{0.49\textwidth}
         \centering
    \includegraphics[width=\textwidth]{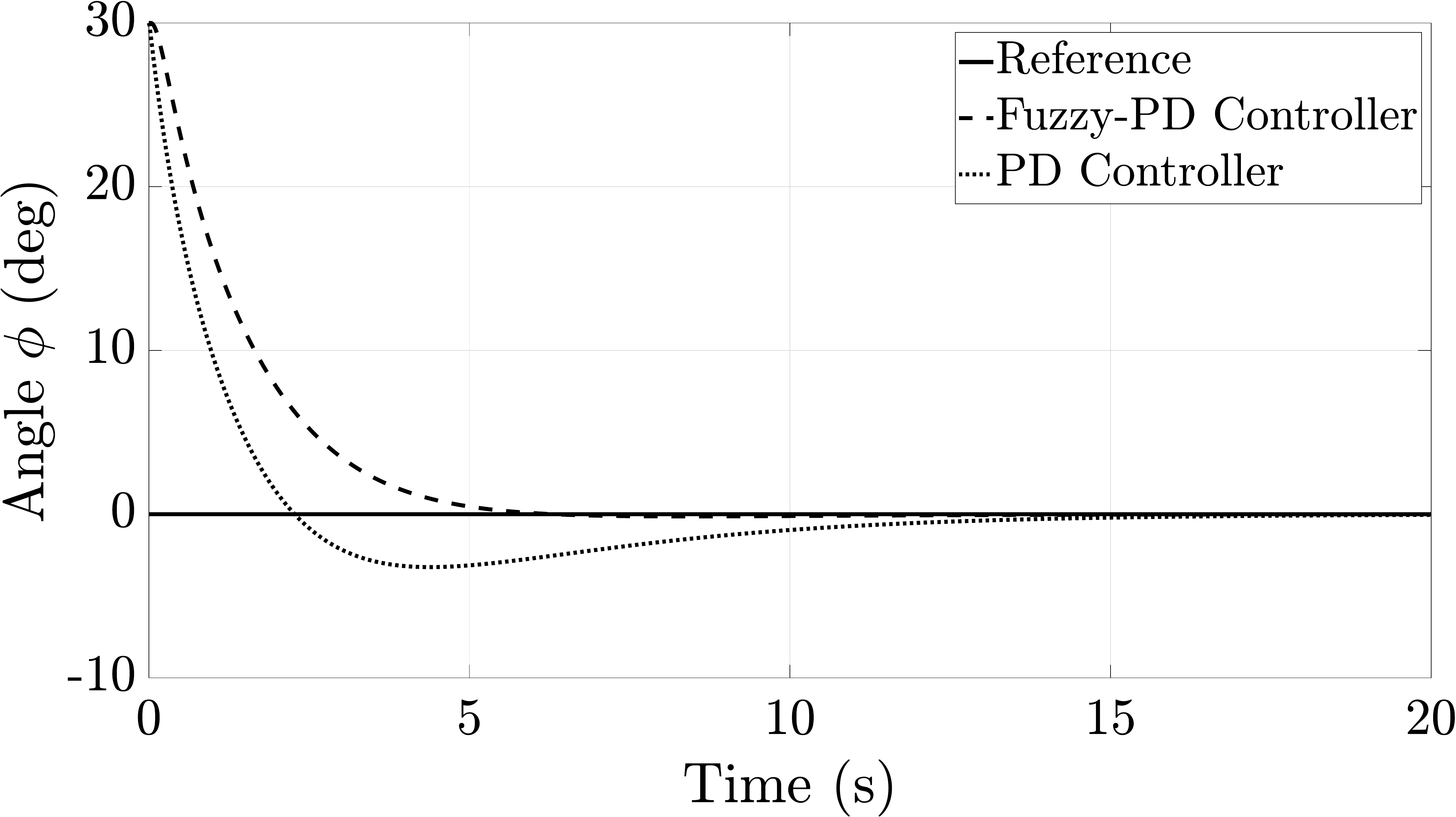}
         \caption{Angles $\phi$.}
     \end{subfigure}
     \begin{subfigure}[b]{0.49\textwidth}
         \centering
    \includegraphics[width=\textwidth]{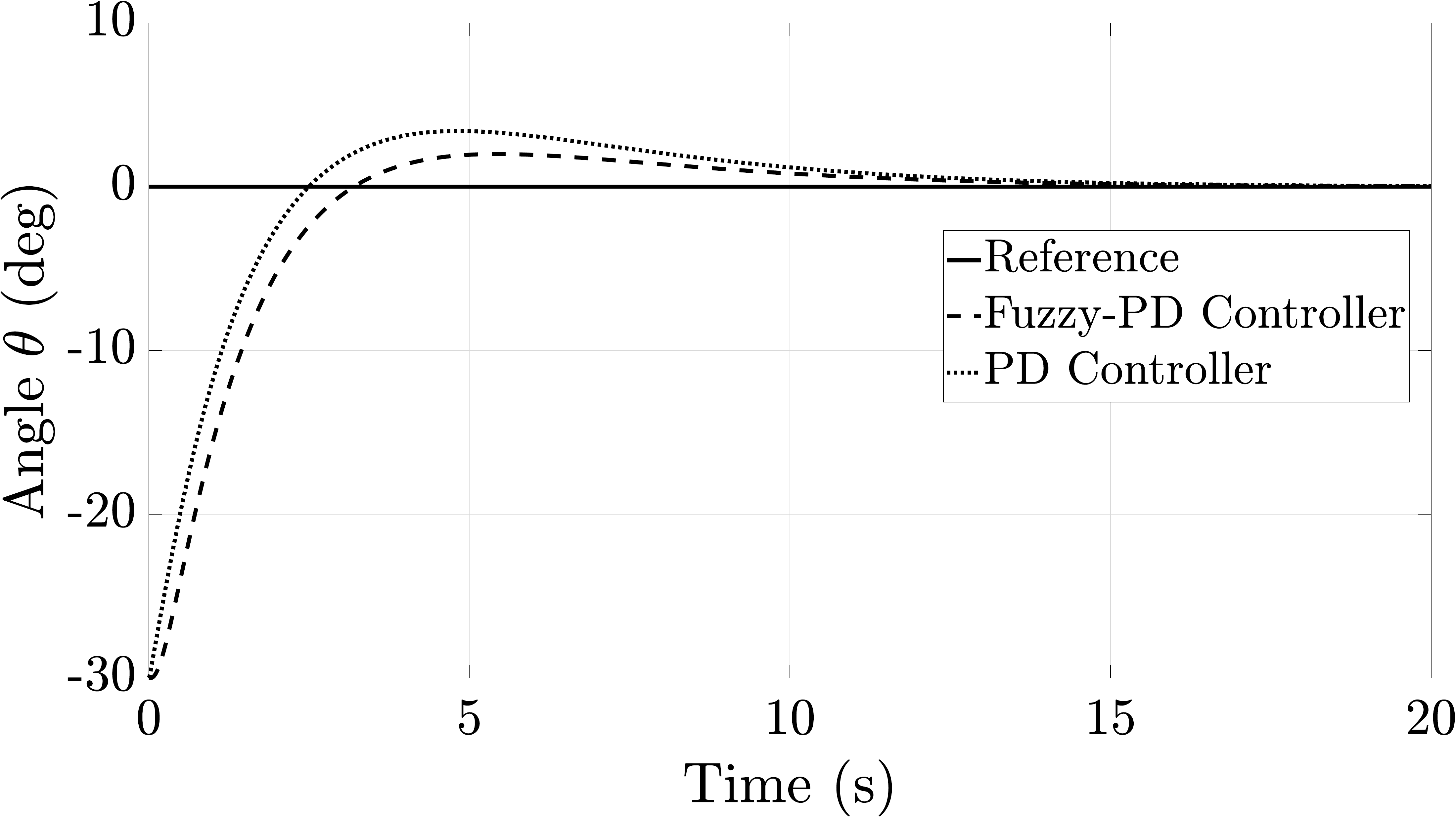}
         \caption{Angle $\theta$.}
     \end{subfigure}
     \begin{subfigure}[b]{0.49\textwidth}
         \centering
    \includegraphics[width=\textwidth]{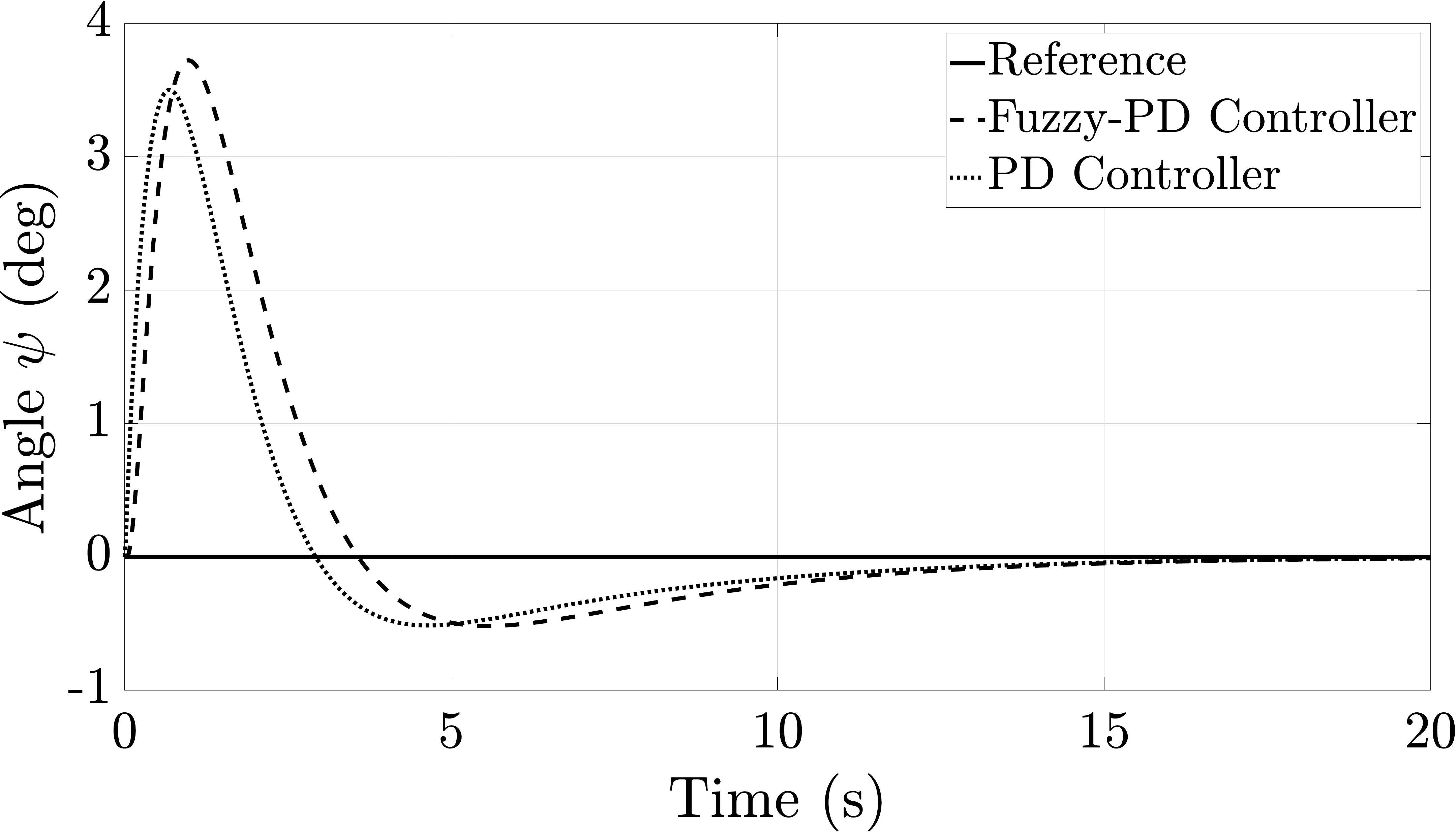}
         \caption{Angle $\psi$.}
     \end{subfigure}
     \begin{subfigure}[b]{0.49\textwidth}
        \includegraphics[width=\textwidth]{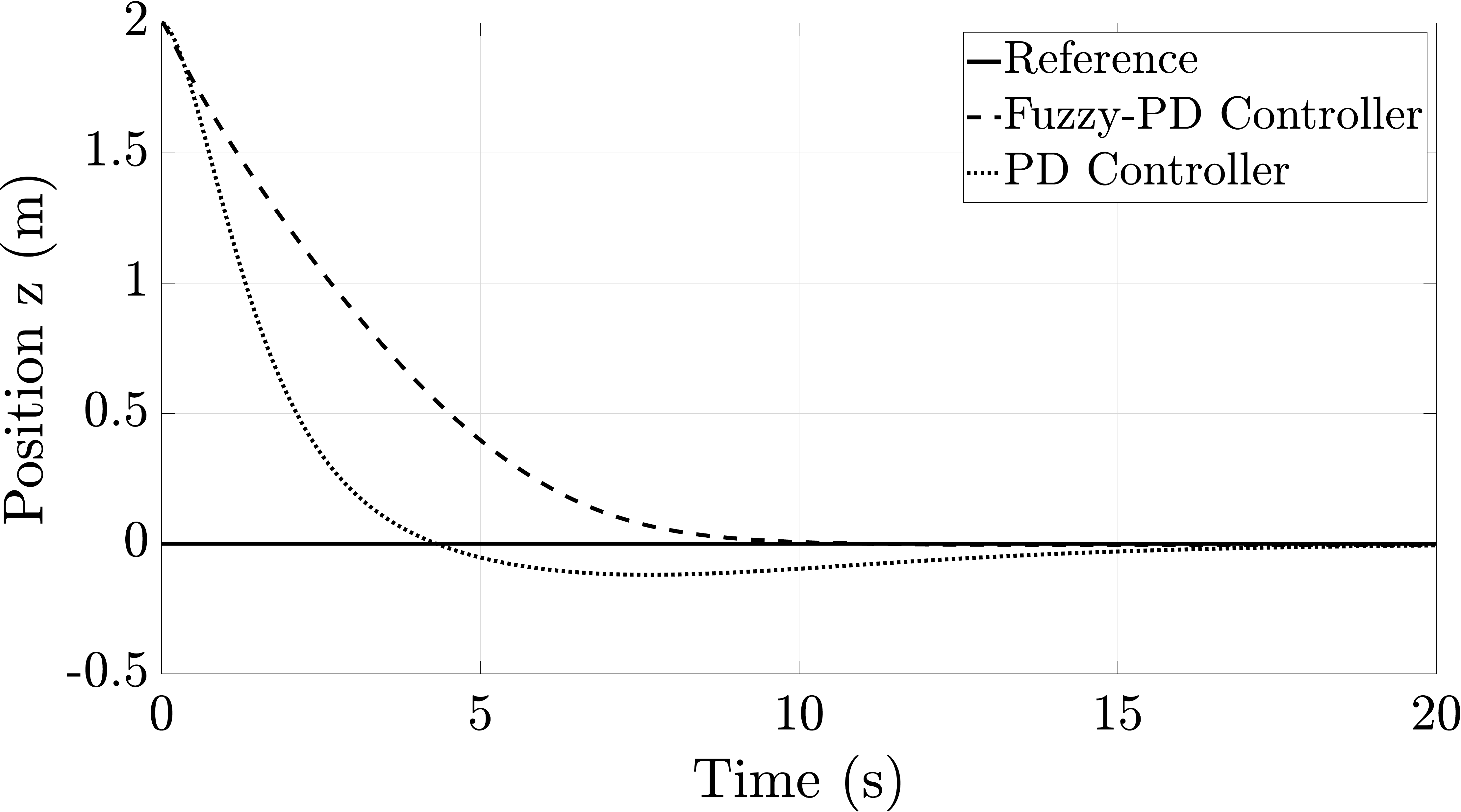}
         \caption{Position z.}
     \end{subfigure}
    \caption{First case response.}
    \label{fig:case1}
\end{figure}

It is obvious from \figurename{ \ref{fig:case1}} that the fuzzy controller is far way better. For the altitude, the settling time of the PD controller and the fuzzy controller is 14 and 8 respectively. The fuzzy controller is 43\% times better than the PD controller. The fuzzy controller reaches the desired state with no overshoot where as the PD controller responses with an overshoot of about 20\%. \\

For the angle $\phi$, the settling time of the PD controller and the fuzzy controller is 12 and 5 respectively. The fuzzy controller is 58\% times better than the PD controller. The fuzzy controller reaches the desired state with no overshoot where as the PD controller responses with an overshoot of about 3\%. For the angles $\theta$, no obvious improvements occur.

{}

It is to be noted that the angles $\psi$ deviate a little bit to a maximum of 3.7 deg and then returns to zero. This deviation happens due to the coupling between the dynamics of the quadcopter. For angle $\psi$, this deviation can be ignored.

\subsection{Case II}

For the second case, the initial conditions given to the quadcopter are:
\begin{equation}
    \begin{bnmatrix}
    z_i \\ \phi_i \\ \theta_i \\ \psi_i
    \end{bnmatrix}
    =
    \begin{bnmatrix}
    0 \text{ [m]} \\ 70^\circ\\ -60^\circ\\ 20^\circ
    \end{bnmatrix}
\end{equation}

\begin{figure}[h!]
    \centering
     \begin{subfigure}[b]{0.49\textwidth}
         \centering
    \includegraphics[width=\textwidth]{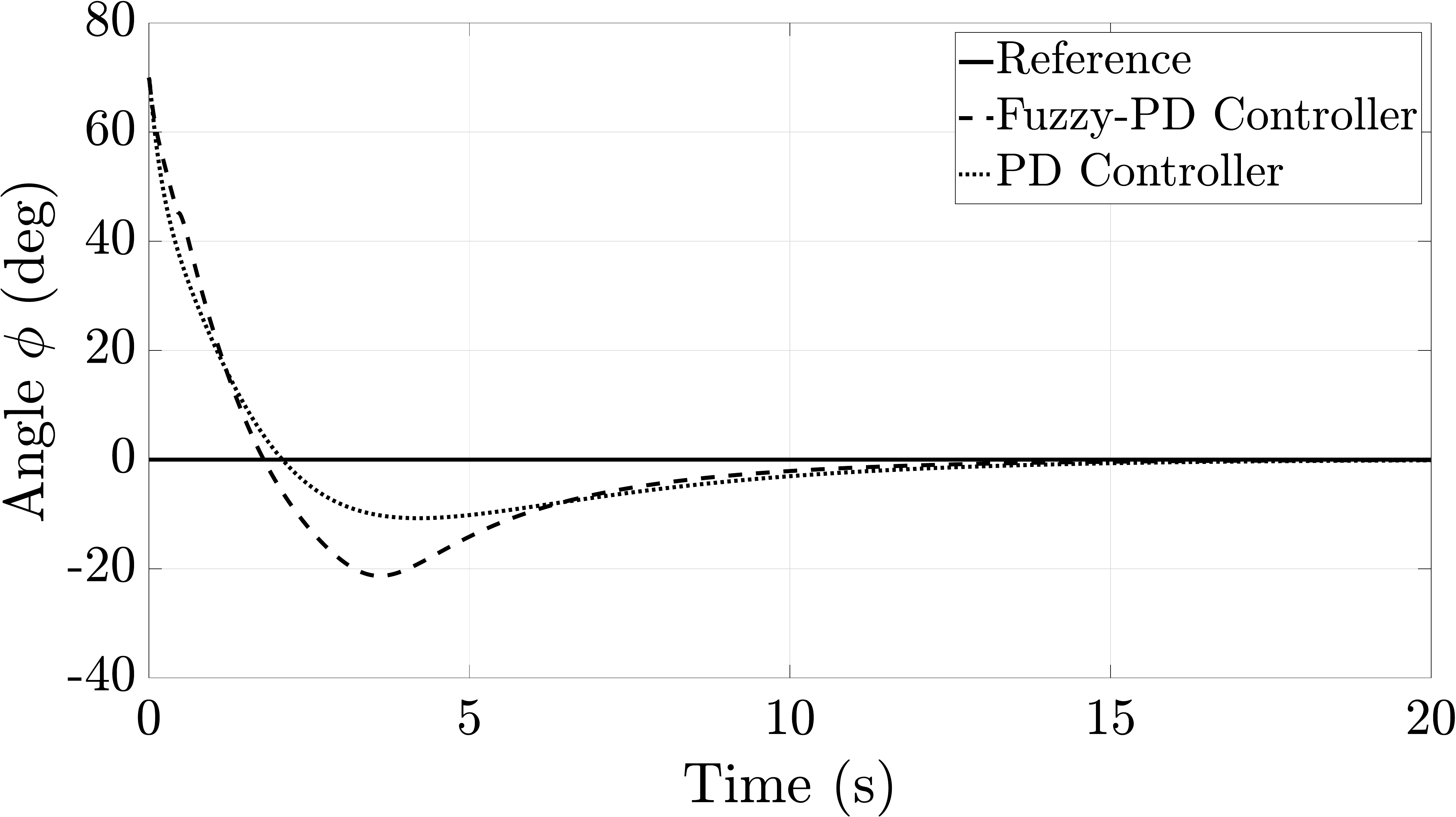}
         \caption{Angles $\phi$.}
     \end{subfigure}
     \begin{subfigure}[b]{0.49\textwidth}
         \centering
    \includegraphics[width=\textwidth]{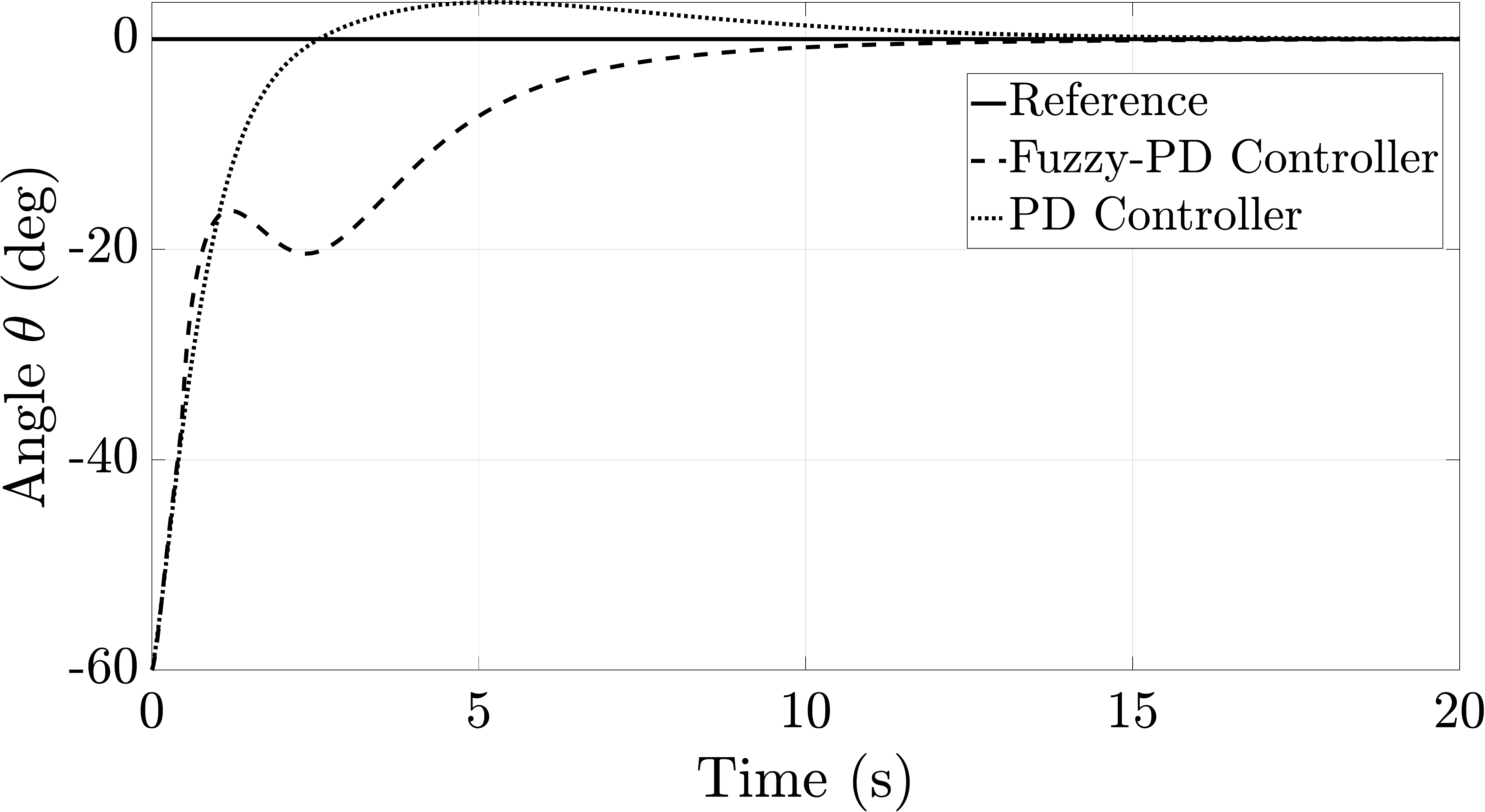}
         \caption{Angle $\theta$.}
     \end{subfigure}
     \begin{subfigure}[b]{0.49\textwidth}
         \centering
    \includegraphics[width=\textwidth]{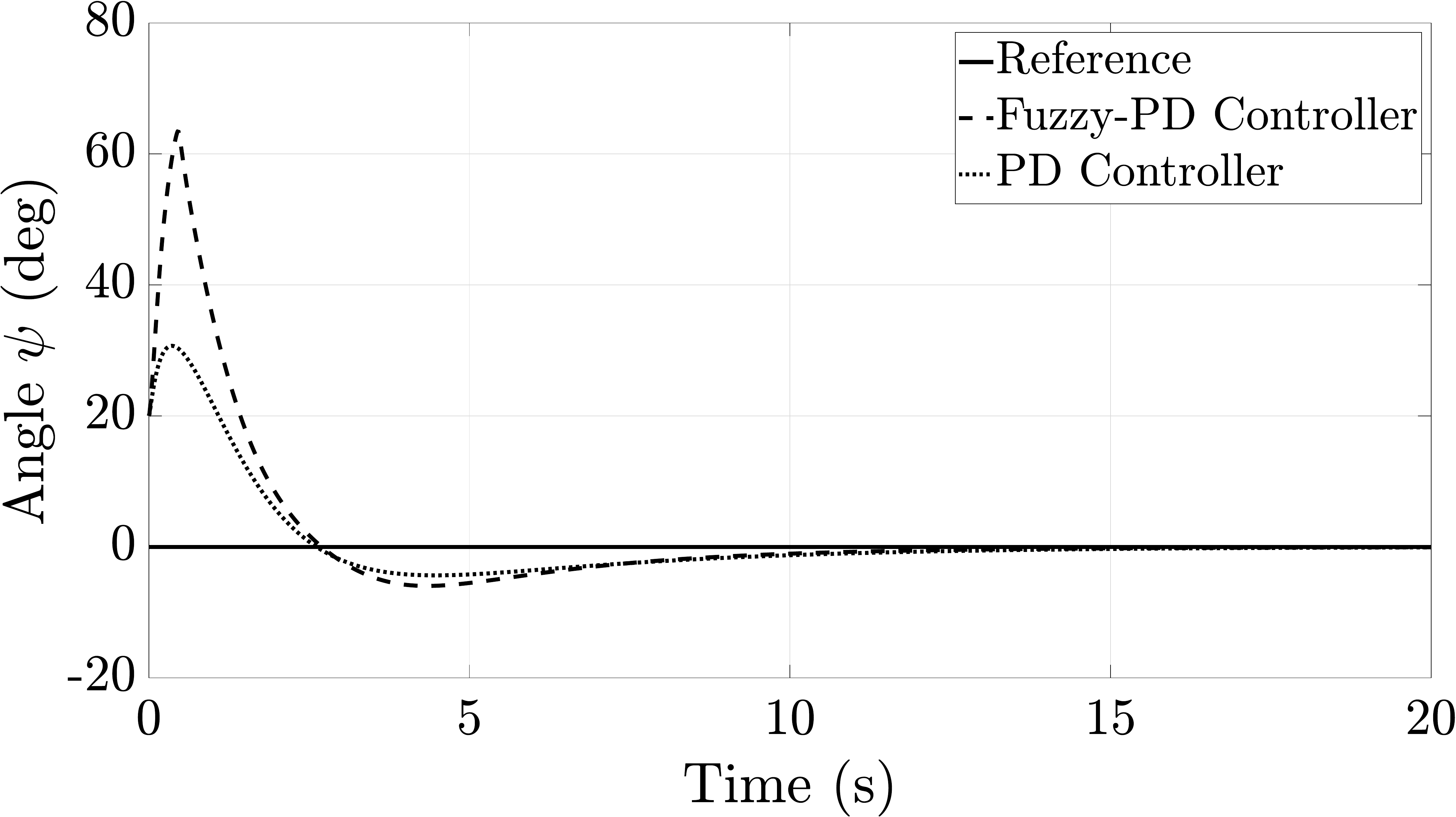}
         \caption{Angle $\psi$.}
     \end{subfigure}
     \begin{subfigure}[b]{0.49\textwidth}
        \includegraphics[width=\textwidth]{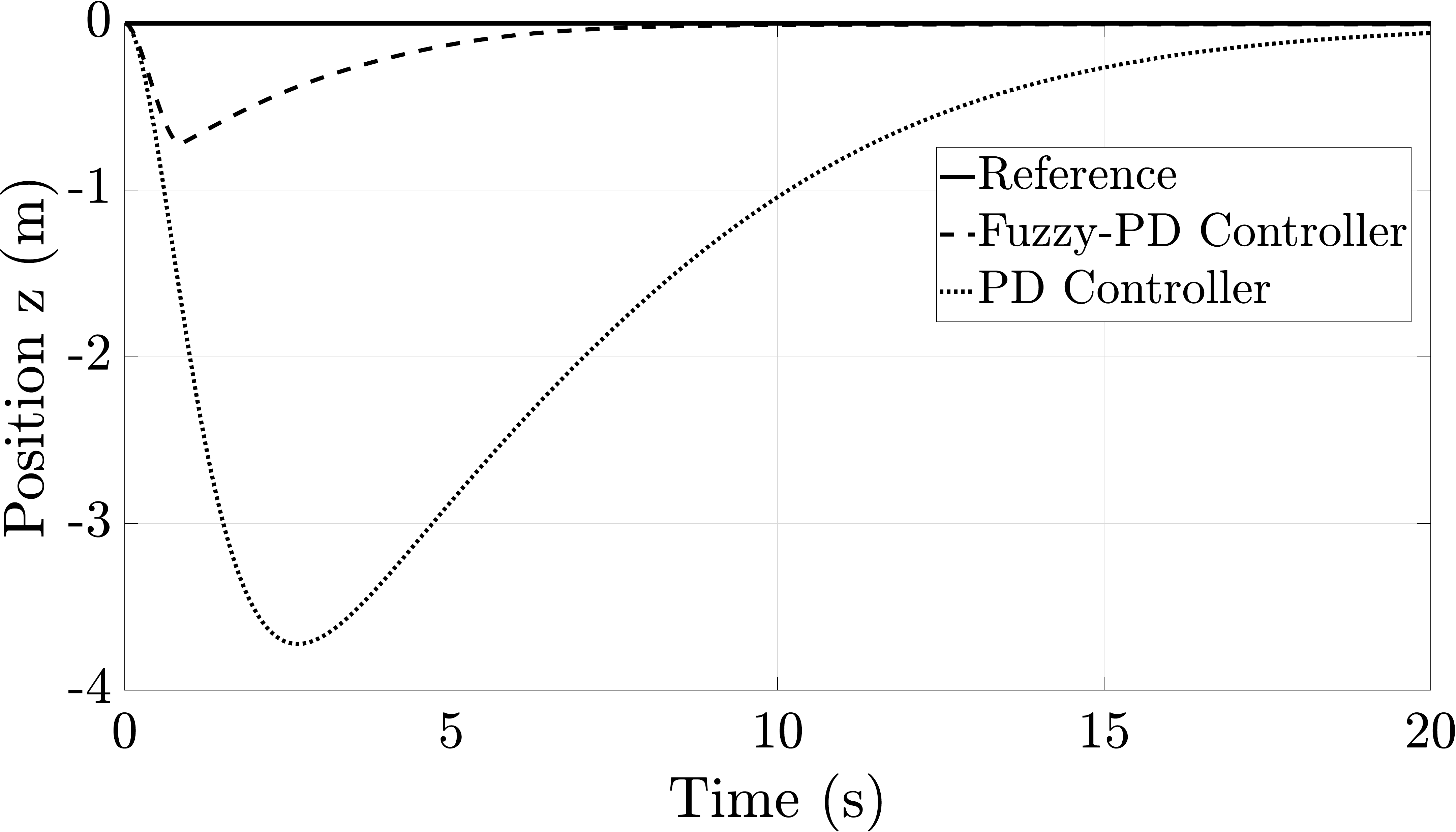}
         \caption{Position z.}
     \end{subfigure}
    \caption{Second case response.}
    \label{fig:case2}
\end{figure}

The response of the attitude is approximately the same as the previous case. The keynote here is for the altitude response. Both controller succeeded in stabilizing the altitude, but the PD controller exhibit large overshoot behaviour which is not acceptable. The fuzzy controller has way less overshoot as noted from \figurename{ \ref{fig:case2}}. \\

It is important to note that even-though the PD controller was developed near the hover state of the quadcopter but it does perform relatively good in stabilizing the quadcopter.

{}

\chapter{Design Testing and Results}
\label{ch7}
\section{System Limitations and Compliance with Design Constraints}
Developing a precise mathematical model of quadcopters can be quite a challenging process due to the environmental impacts on the quadcopter. In this project, we have not considered any external forces that might affect quadcopter dynamics such as aerodynamic forces and drag effects. We also assumed the quadcopter is rigid, symmetrical and the center of gravity is exactly located in the middle, which is not always the case. \\

In addition, we linearized the model and made it only operate in a small state, the hover state. Then, we developed a controller based on this model which makes the controller operate perfectly only in the hover state. We are also constrained by the physical properties of motors where the control signal should be feasible.  

\section{Solution Impact}
Robots have been used in many different areas throughout human history, including in many of our most important and significant technological achievements. From building the pyramids to operating our satellites, robots have been crucial in the development of the modern world. One interesting area is the use of drones as aerial surveillance tools. Today, drones are being used for a wide variety of applications. Some have been used to map out parts of the world quickly and accurately, while others have been used to deliver packages to remote locations, and still others have been used to protect the country's borders from illegal crossings.\\ 

Drones are also being used as tools to help in the search for survivors after natural disasters, and in the search for missing people. They are also being used as part of the defense of Earth's natural resources and in the fight against disease-spreaders, drug smugglers, and terrorists. Drones are one of the newest and most important tools of warfare, and their presence and impact on our society are only going to grow as the technologies in the drone industry continue to evolve.\\ 

They will soon be seen everywhere and will be used as tools in every aspect of our lives, so much so that we may lose sight of their true role in our lives. Drones are not just a tool for war and disaster relief. They are tools for the 21st century. They are a platform for innovation and collaboration. They can provide real world assistance to people in need in the midst of natural disasters and other catastrophes. Drones can assist with relief efforts by providing low cost and high precision surveillance of the disaster area, enabling people to make life-saving decisions and saving them time and money. Drones can be used to provide communications and humanitarian aid. They can also be used as a platform to enhance security, allowing for a better understanding of the terrain, and protecting people. They can be used to assist in search and rescue missions. And they can be used to protect property and prevent crimes.\\ 

But the use of drones has also been used for more nefarious purposes. One such use was the widespread use of Predator drones to kill suspected terrorists. The use of drones for assassination has raised many privacy and ethical concerns and there are even calls for a ban on the use of drones. But as more countries begin using drones to spy on their citizens and spy on other countries, it's important to be cautious when it comes to how drones are used.



\chapter{Conclusion and Future work}
\label{ch8}
\section{Conclusion}
In this project, we developed a simple mathematical model of the quadcopter in a step by step manner. The derived model is then implemented in MATLAB Simulink 2019a, a non-linear simulation environment. To validate the model, we simulated the response of the open-loop system. \\

Now that the model is validated, we designed a PID controller to stabilize the quadcopter and for that we used a linearized version of the quadcopter model. The linearization is achieved around a given operation point, which is the hover state, that made the calculations more simple and decoupled the system. \\

We then implemented the developed PID controller and simulated the closed-loop response of the system. As expected, we found that the model is working properly and gives reasonable results. We then went further and expanded the region of operation of the quadcopter by developing a non-linear controller, the ANFIS controller. We showed that the ANFIS controller clearly outperforms the classical PID controller when the quadcopter operates far from the hover state. \\

Finally, we successfully deployed the developed controllers on hardware but due to Covid-19 pandemic, we were unable to test and debug. Therefore, we have not included any details regarding the deployment process.
\section{Recommendations for Future Work}
This project can be expanded to include many new features or even use it to try any new control algorithms. We can recommend a few improvements as follows:
\begin{enumerate}
    \item Develop, implement and simulate aggressive controllers.
    \item Derive the equations that govern the positions of the quadcopter. Then, build and implement trajectory tracking control loop.
    \item Improve the current mathematical model of the quadcopter to include more environmental effects.
    \item Deploy, debug and test custom made trajectory tracking controllers based on GPS readings.
\end{enumerate}
\appendix
\addtocontents{toc}{\protect\contentsline{chapter}{Appendices}{}{chapter*.\thepage}}
\chapter{Parameter determination}
\label{app1}
This chapter summarizes the process of measuring the physical properties of the quadcopter.

\section{Mass}
\begin{figure}[h!]
    \centering
     \begin{subfigure}[b]{0.49\textwidth}
         \centering
        \includegraphics[width=\textwidth]{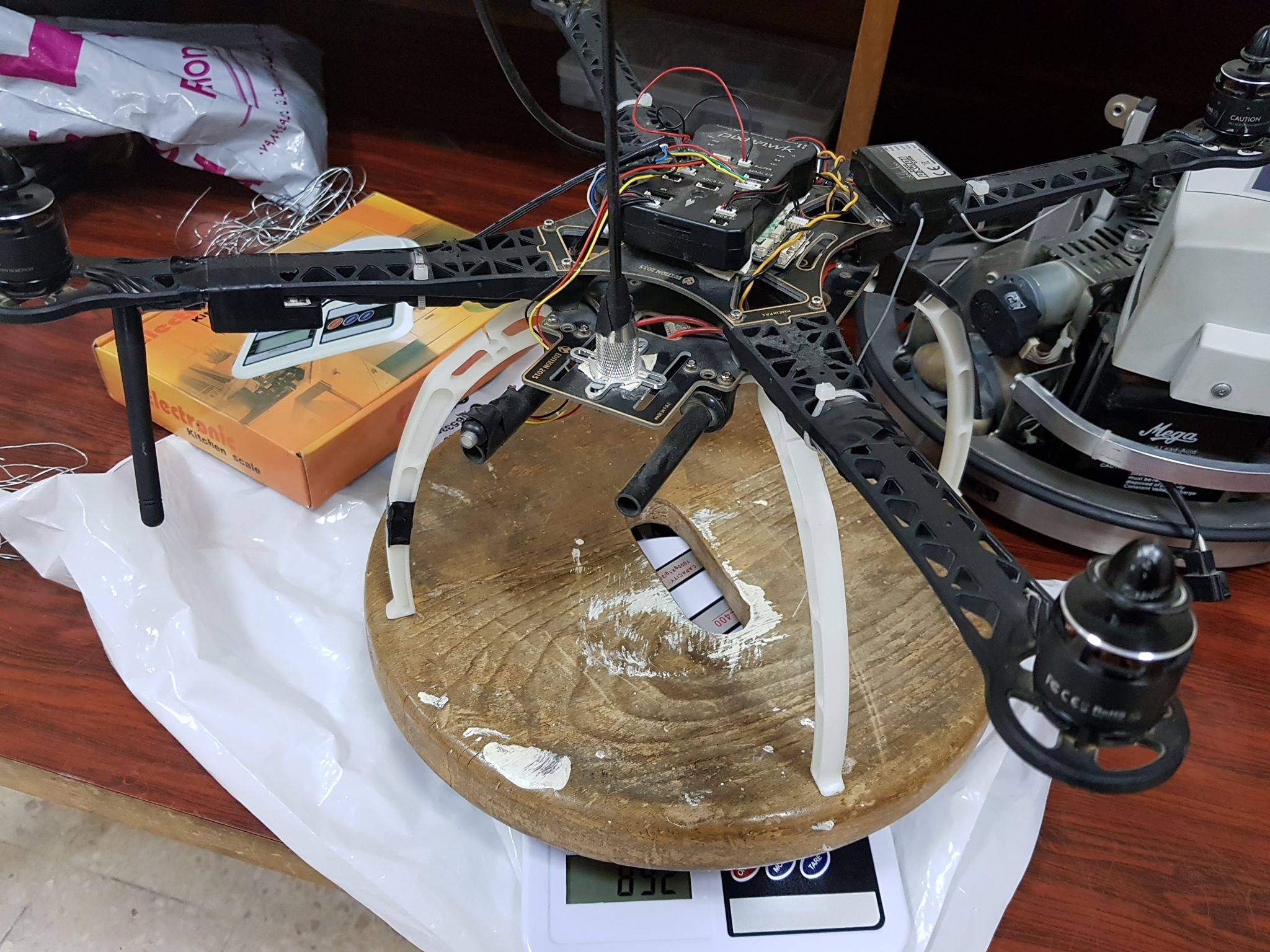}
     \end{subfigure}
     \begin{subfigure}[b]{0.49\textwidth}
         \centering
        \includegraphics[width=\textwidth]{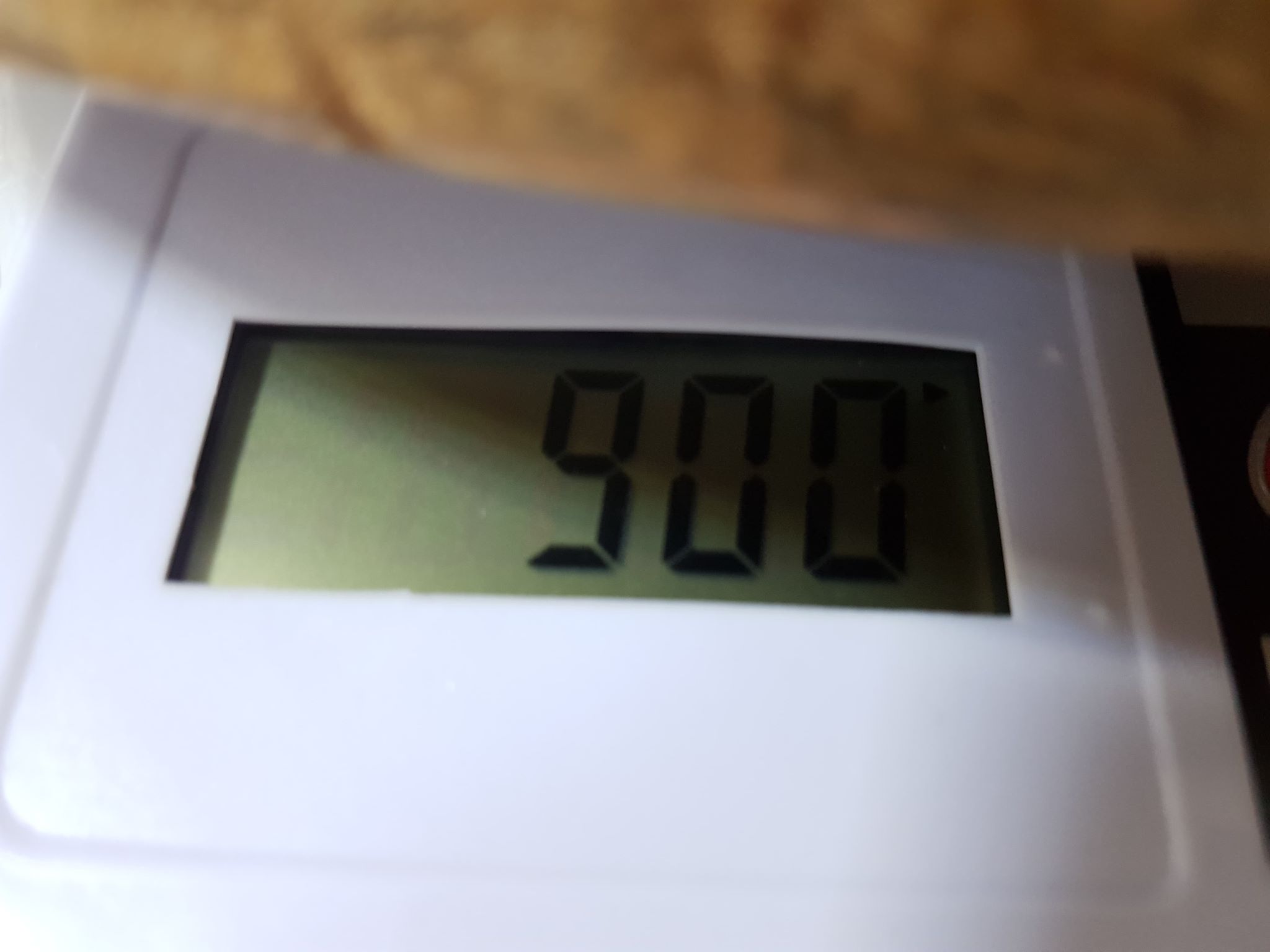}
     \end{subfigure}
    \caption{Mass measurement.}
    \label{fig:mass}
\end{figure}

The quadcopter's mass was measured by a food scale, as shown in \figurename{ \ref{fig:mass}}. A flat object was placed on the scale to ensure that the quadcopter remained stable on the scale. The quadcopter's mass, $m$, was equal to 900 grams.

{}

\section{Arm length}
\begin{figure}[h!]
    \centering
    \includegraphics[width=0.9\textwidth]{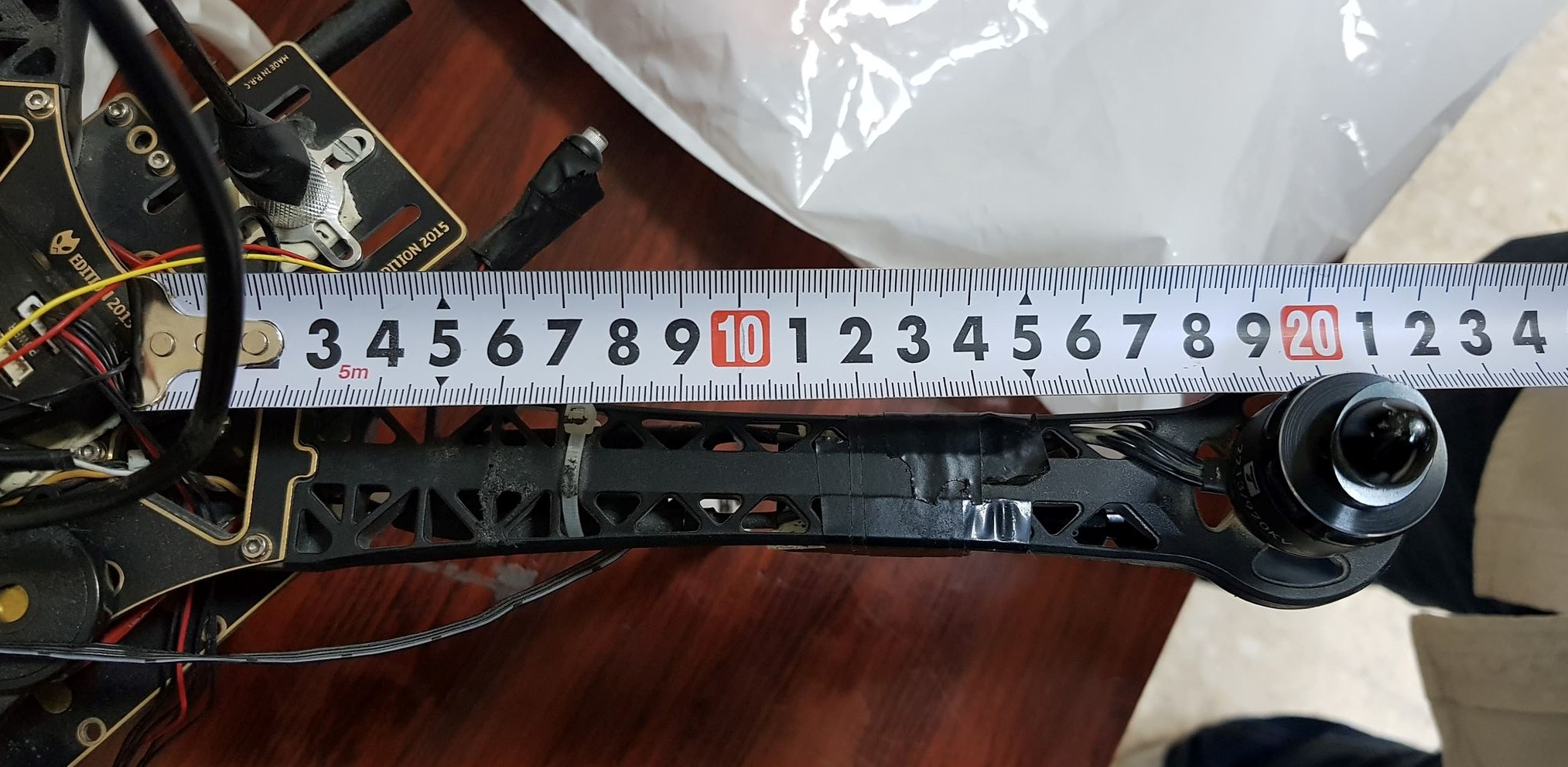}
    \caption{Arm length measurement.}
    \label{fig:arm}
\end{figure}

The arm length was measured using a meter as shown in \figurename{ \ref{fig:arm}}, and it was found to be 22 cm.

{}

\section{Moment of inertia}
The process of measuring the moment of inertia of the quadcopter depends on an experiment known as the bifilar pendulum \cite{amadi2018design}. In this experiment, the quadcopter is suspended using two strings of equal length. \\

Once the quadcopter is suspended and stops moving, the quadcopter is shifted slightly from the equilibrium position. A stopwatch starts in conjunction with the the start of the quadcopter's movement and the number of seconds it takes for the quadcopter to complete ten oscillations is recorded. This procedure is repeated two more times in order to calculate the mean. \\

\begin{figure}[H]
    \centering
     \begin{subfigure}[b]{0.3\textwidth}
         \centering
        \includegraphics[width=0.9\textwidth]{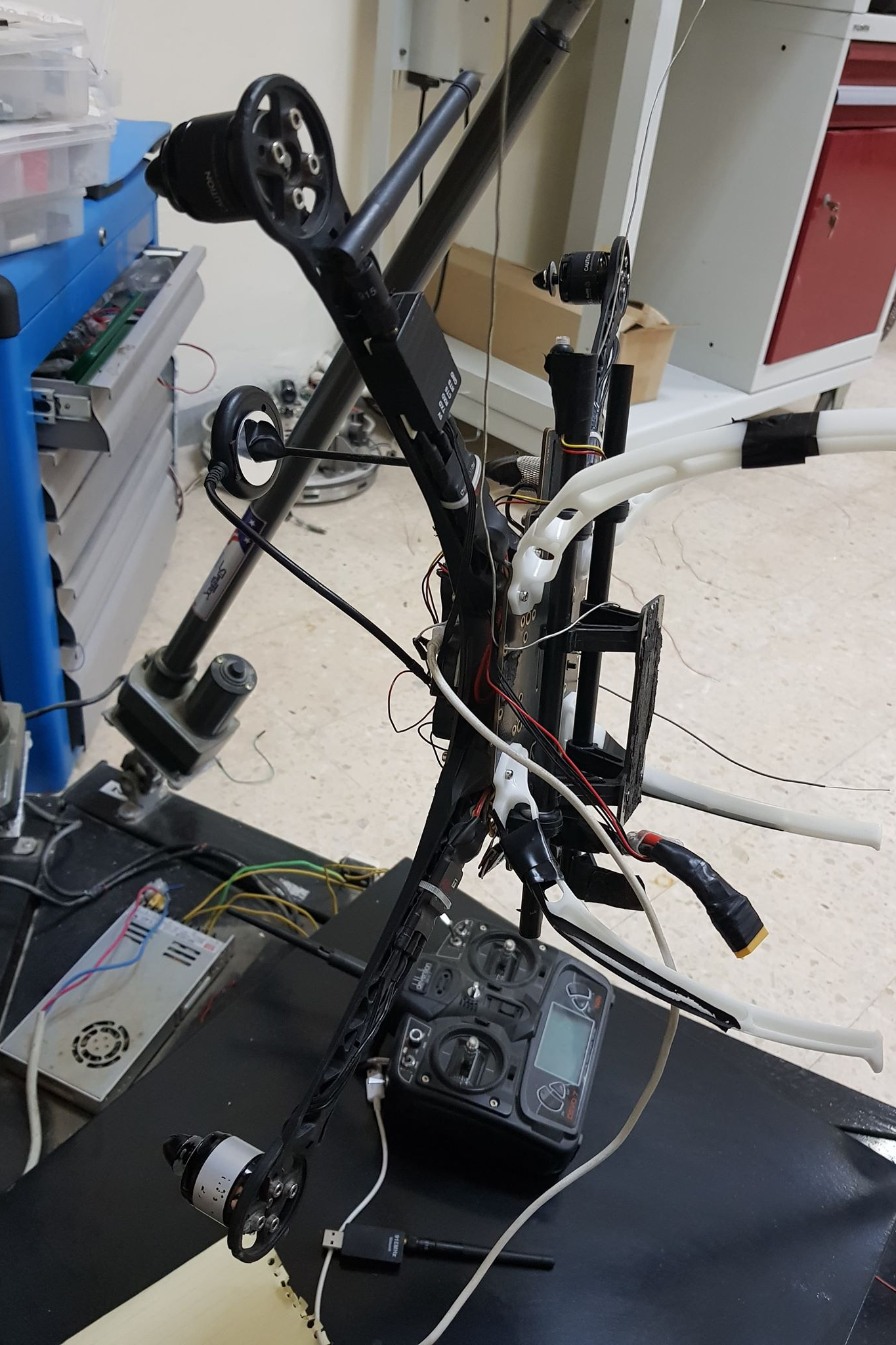}
     \end{subfigure}
     \begin{subfigure}[b]{0.3\textwidth}
         \centering
        \includegraphics[width=0.9\textwidth]{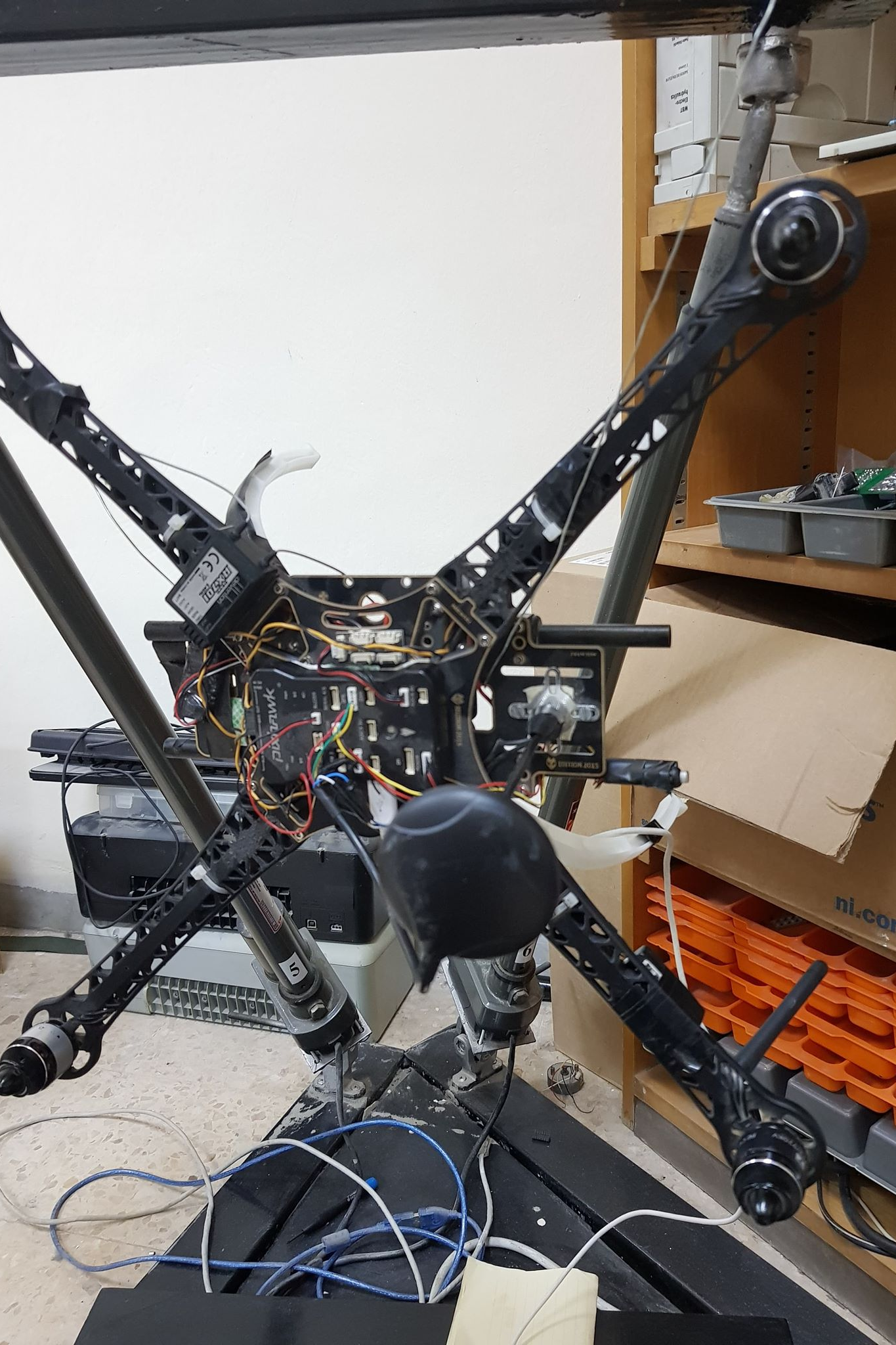}
     \end{subfigure}
     \begin{subfigure}[b]{0.3\textwidth}
         \centering
        \includegraphics[width=0.9\textwidth]{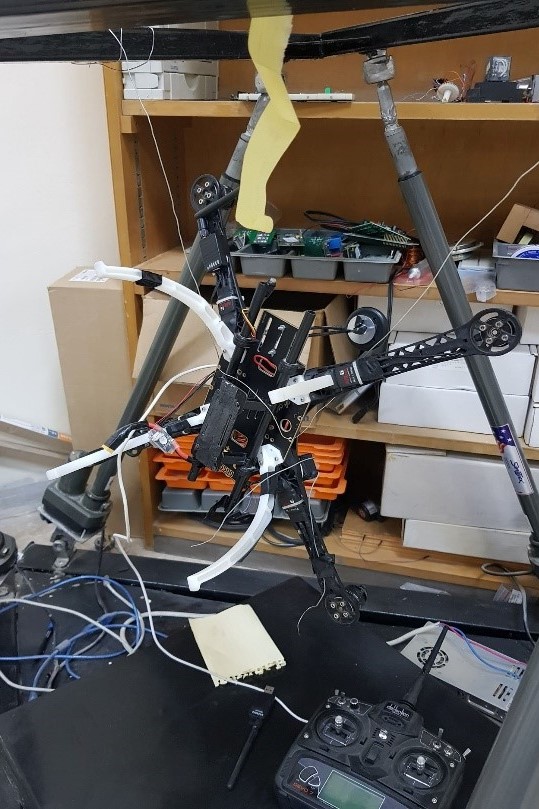}
     \end{subfigure}
    \caption{Moment of inertia measurements.}
    \label{fig:i}
\end{figure}

This procedure is repeated in the other two axes as presented in \figurename{ \ref{fig:i}}. The results from the experiments are presented in Table \ref{tab:paramI} where we recorded the results three times for each axis and took the average. \\

{}

\begin{table}[H]
    \centering
    \caption{Results for the moment of inertia experiment.}
    \label{tab:paramI}
    \setlength{\extrarowheight}{.95ex}
    
    \begin{tabular}{|>{\columncolor{Gray}}c|c|c|c|}
        \toprule
        \rowcolor{Gray}
         \textbf{Time (s)} & \textbf{About x}  &  \textbf{About x} &  \textbf{About z}  \\
         \midrule 
         T$_1$ & 12.15 & 12.76 & 13.32 \\
         T$_2$ & 12.23 & 13.15 & 13.12 \\
         T$_3$ & 12.15 & 13.09 & 13.41 \\
         \bottomrule
    \end{tabular}
\end{table}{}

After the measurements are determined, the moment of inertia is calculated for each axis from the following equation \cite{amadi2018design}:
\begin{equation}
    I = \frac{mgT^2d^2}{16 \pi^2 L}
\end{equation}
where,\\
$m$ - mass of the quadcopter, kg\\
$g$ - gravity acceleration, ms$^{-2}$\\
$T$ - period of one oscillation, s\\
$d$ - distance between strings, m\\
$L$ - length of string, m\\

\par For this experimental setup the following values were measured,\\
$L = 0.24$ m \\
$d=0.21$ m \\

\chapter{Linear mapping proof}
\label{app2}

In order to transform a given function, $f$, to the $s$ domain as a transfer function, it has to satisfy the linear mapping conditions, additivity and homogeneity \cite{whitelaw2019introduction}. Additivity is the operation of addition while homogeneity is the operation of scalar multiplication. These two conditions are:

\begin{align}
    f(u+v) &= f(u)+f(v) \\[10pt]
    f(cv) &=  cf(v)
\end{align}

Recall, the linearized altitude equation is as follows:
\begin{equation}
    \ddot{z} = \frac{F}{m} - g
\end{equation}

After reassigning variables, the equation becomes:
\begin{equation}
    f(F) = \frac{F}{m} - g
\end{equation}

Checking for the additivity condition:
\begin{equation}
    \begin{split}
        f(u) &= \frac{u}{m} - g, \\[10pt]
        f(v) &= \frac{v}{m} - g, \\[10pt]
        f(u) + f(v) &= \frac{u+v}{m} - 2g,\\[10pt]
        f(u+v) &= \frac{u+v}{m} - g, \\[10pt]
        f(u) + f(v) &\neq f(u+v)
    \end{split}
\end{equation}

Checking for the homogeneity condition:
\begin{equation}
    \begin{split}
        f(cv) &= \frac{cv}{m} - g, \\[10pt]
        cf(v) &= \frac{cv}{m} - cg, \\[10pt]
        f(cv) &\neq cf(v)
    \end{split}
\end{equation}

This proves that the altitude equation does not satisfy the linear mapping conditions, therefore it cannot be directly transformed to transfer function.

\chapter{Steady state error proof}
\label{app3}

According to the final value theorem, the steady state behaviour of $f(t)$ when $t\to \infty$ is related to the behaviour of $sF(s)$ when $s\to 0$ provided that $f(t)$ is the limit of $f(t)$ as $t\to \infty$ exists \cite{franklin2015feedback}. This means:
\begin{equation}
    \lim_{t\to\infty} f(t) = \lim_{s\to0} sF(s)
\end{equation}

Recall, the second order system is:
\begin{equation}
    G(s) = \frac{F(s)}{R(s)} = \frac{1}{Ks^2}
\end{equation}
where $K$ is a constant and $R(s)$ is the input. \\

The error of the system, $E(s)$, is found to be \cite{shinners1998modern}:
\begin{equation}
    E(s) = \frac{R(s)}{1+G(s)}
\end{equation}

For a step input, $1/s$, the error becomes:
\begin{equation}
    \begin{split}
        e(\infty) &= \lim_{s\to0} sE(s) \\
        &= \lim_{s\to0} s\frac{1/s}{1+G(s)} \\
        &= \lim_{s\to0} \frac{1}{1+G(s)} \\
        &= \lim_{s\to0} \frac{1}{1+\frac{1}{Ks^2}}  \\
        &= 0
    \end{split}
\end{equation}
where $e(t)$ is the error in time domain. \\

For a ramp input, $1/s^2$, the error becomes:
\begin{equation}
    \begin{split}
        e(\infty) &= \lim_{s\to0} sE(s) \\
        &= \lim_{s\to0} s\frac{1/s^2}{1+G(s)} \\
        &= \lim_{s\to0} \frac{1/s}{1+G(s)} \\
        &= \lim_{s\to0} \frac{1/s}{1+\frac{1}{Ks^2}}  \\
        &= \lim_{s\to0} \frac{s}{s^2+\frac{1}{K}} \\
        &= 0
    \end{split}
\end{equation}

This proves that the steady state error for the system dynamic equations is zero due to step and ramp inputs.

\bibliographystyle{unsrtnat}
\bibliography{main}
\end{document}